\documentclass[runningheads]{llncs}
\usepackage[T1]{fontenc}
\usepackage{graphicx}
\usepackage{amsmath}
\usepackage{multirow}
\usepackage{adjustbox}
\usepackage{subcaption}
\usepackage{array}
\usepackage[table]{xcolor}
\usepackage{amssymb}
\usepackage{calc}
\usepackage[normalem]{ulem}
\usepackage{comment}
\usepackage{enumitem}

\begin{document}

\title{%CHROMA: Leveraging Inter-Channel Color-Space Correlations for AI-Generated Image Detection
%CHROMA: Color Correlation Analysis for AI-Generated Image Detection
CHROMA: Detecting AI-Generated Images through Inter-Channel Color-Space Correlations\thanks{This work was partially supported by Agencia Nacional de Investigación e Innovación (ANII, Uruguay) grant POS\_NAC\_2023\_1\_178226. The experiments presented in this paper were carried out using ClusterUY (https://cluster.uy).}
}
%
%\titlerunning{Abbreviated paper title}
% If the paper title is too long for the running head, you can set
% an abbreviated paper title here
%
\author{Juan Pablo Sotelo\inst{1}\orcidID{0009-0005-2070-8529} \and
Marina Gardella\inst{2}\orcidID{0000-0003-2465-2014} \and
Pablo Musé\inst{1,2}\orcidID{0000-0002-9199-6469}}
\authorrunning{J. Sotelo et al.}
% First names are abbreviated in the running head.
% If there are more than two authors, 'et al.' is used.
%
\institute{
 Instituto de Ingeniería Eléctrica, Facultad de Ingeniería, Universidad de
la República, Montevideo, Uruguay \\
\email{\{juan.pablo.sotelo.silva, pmuse\}@fing.edu.uy} \and
Université Paris-Saclay, ENS Paris-Saclay, CNRS, Centre Borelli, Gif-sur-Yvette, 91190 France
\email{marina.gardella@ens-paris-saclay.fr}}
\maketitle
\begin{abstract}
The rapid adoption of diffusion and large-scale generative models has made it increasingly challenging to distinguish synthetic %and manipulated 
imagery from real photographs. While automated detectors have been proposed, 
their generalization to unseen generators remains brittle. To address this limitation, we investigate \emph{inter-channel color correlations}, a lightweight and underexploited forensic cue.
We first demonstrate that LPIPS, a widely used perceptual metric, exhibits inconsistent responses to perturbations that selectively alter channel dependence across different color-space parameterizations, indicating that cross-channel statistics are not uniformly constrained by common perceptual training objectives. Motivated by this, we analyze the distributions of pairwise inter-channel correlation features across multiple color spaces.
Our analysis reveals systematic, generator-specific differences in these distributions, with RGB and Lab color spaces providing the most apparent separation between real and generated images. Building on this, 
we introduce \textsc{Chroma}, a detector of AI-generated images which augments standard RGB inputs with inter-channel correlation maps and employs a fixed CNN backbone trained with a modest computational budget. We assess its robustness under both single-generator training and a limited multi-generator supervision regime, where only a few samples from additional generators are available. Across a standard benchmark protocol, correlation-augmented inputs improve real-vs-generated discrimination and robustness, yielding performance competitive with recent detectors while maintaining a simple architecture and training procedure.
\\Code is available at \url{https://github.com/JPSoteloSilva/CHROMA}.

%\vspace*{-1.2em}
\end{abstract}

\newlength{\toprowH}
\setlength{\toprowH}{0.18\textheight} % tweak if needed

\newcommand{\topimg}[1]{%
  \includegraphics[width=\linewidth,height=\toprowH,keepaspectratio]{#1}%
}

\begin{figure*}[t]
    \centering
    \begin{subfigure}[t]{0.32\textwidth}
        \centering
        \includegraphics[width=\linewidth]{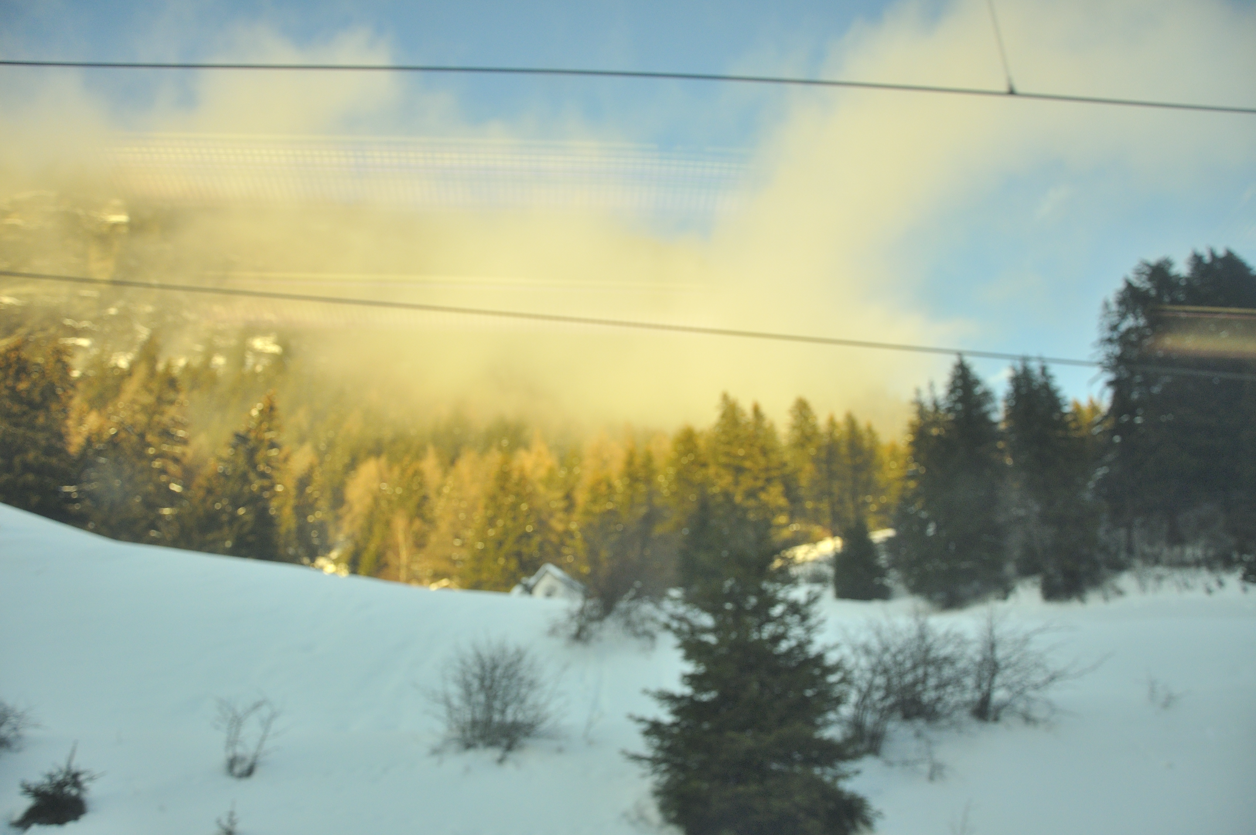}
        \caption{\scriptsize Real}
        \label{fig:gpt_real}
    \end{subfigure}\hfill
    \begin{subfigure}[t]{0.32\textwidth}
        \centering
        \includegraphics[width=\linewidth]{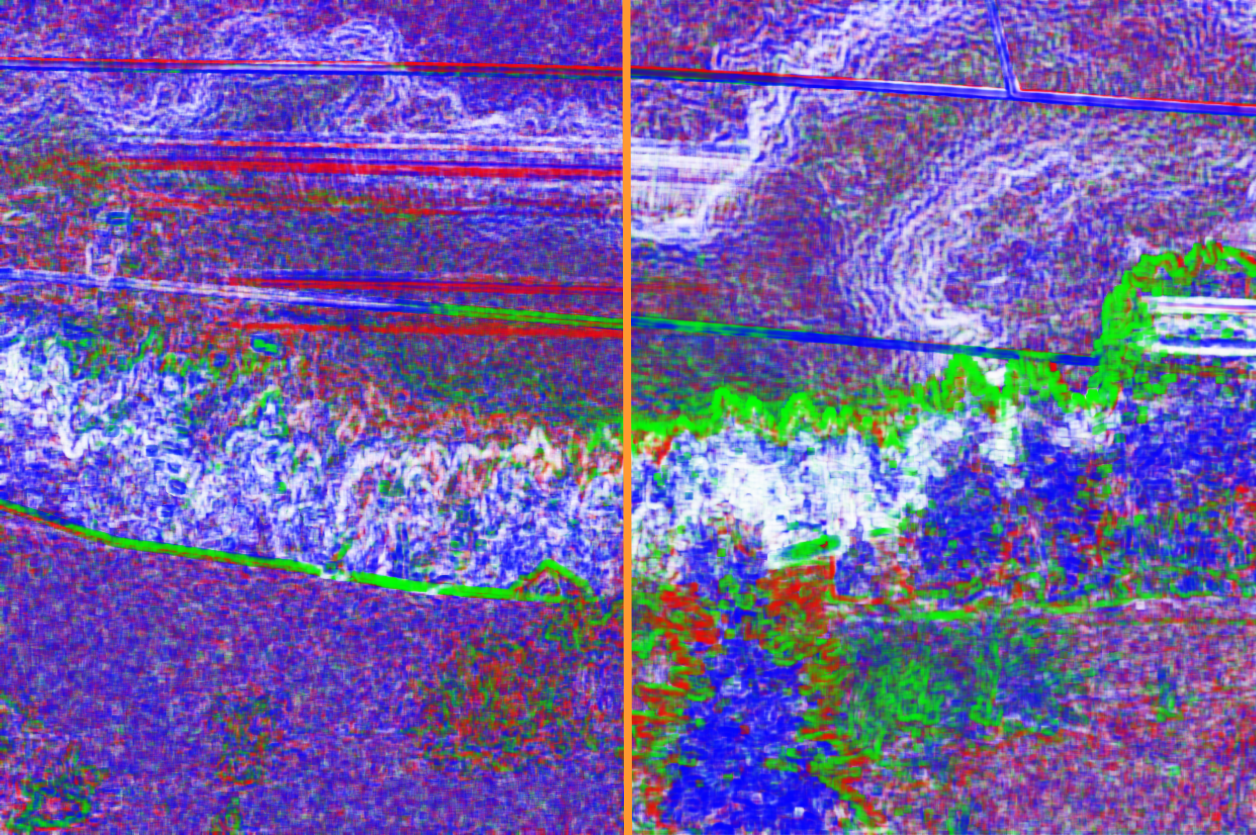}
        \caption{\scriptsize Lab correlation (real/gen.)}
        \label{fig:gpt_corr}
    \end{subfigure}\hfill
    \begin{subfigure}[t]{0.32\textwidth}
        \centering
        \includegraphics[width=\linewidth]{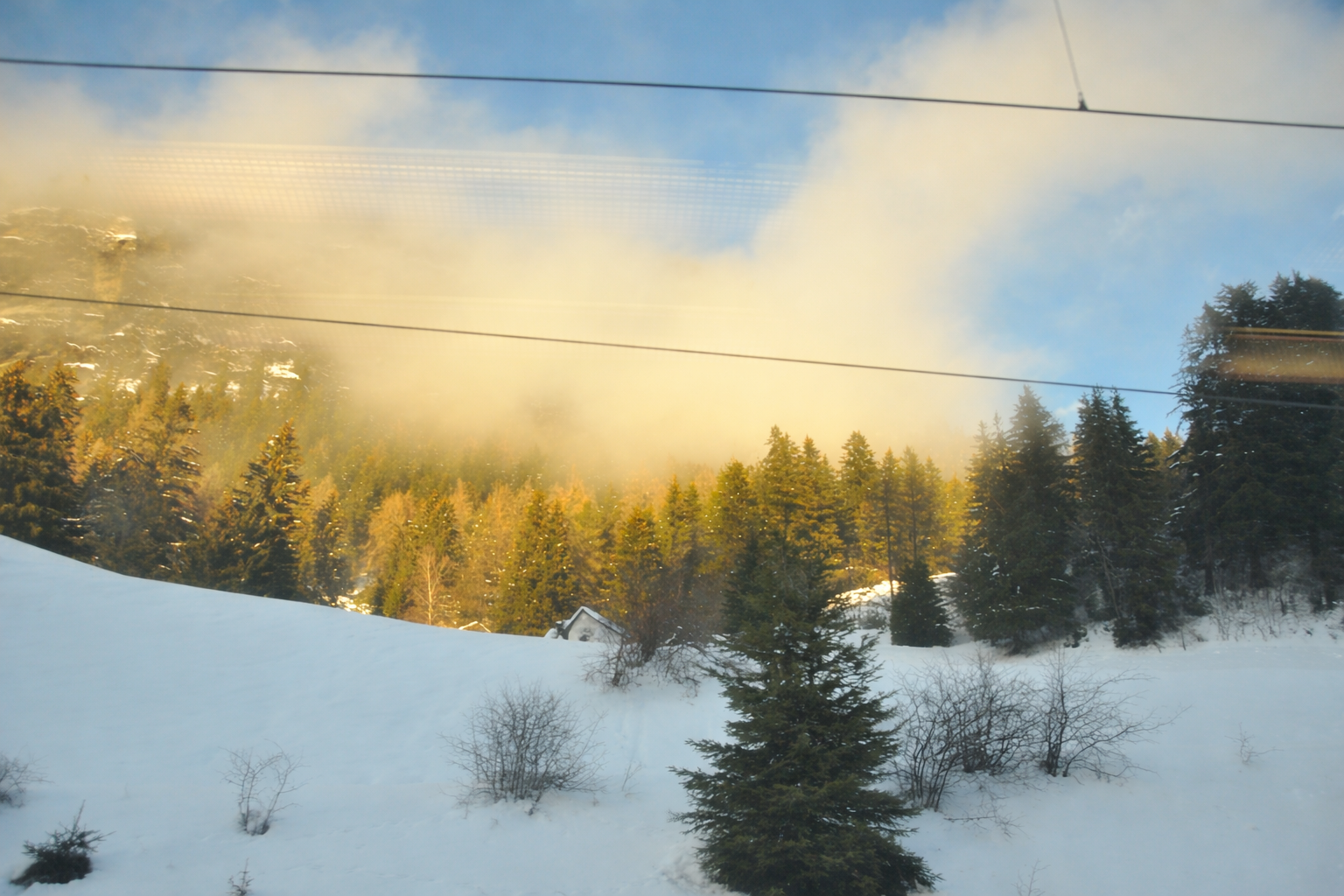}
        \caption{\scriptsize Generated}
        \label{fig:gpt_fake}
    \end{subfigure}

    \caption{\textbf{Color-correlation cues.}
    RAISE-1K real image~\cite{dangnguyen2015raise} vs.\ a visually matched GPT-Image~1 replica~\cite{openai_gptimage1}; the split $\mathrm{Lab}$ correlation map exposes structured chromatic and texture differences under matched content.}
    \label{fig:cover_chroma}
    \vspace*{-0.8em}
\end{figure*}

\section{Introduction}
\label{sec:introduction}

The ability to generate photorealistic images from text prompts %or to seamlessly edit existing photographs 
has advanced rapidly, driven by diffusion models and large-scale generative training~\cite{DMbeatGAN}. These systems are now widely used in creative workflows and deployed in consumer-facing applications, enabling the low-cost creation of synthetic %or manipulated 
imagery at scale. This raises pressing concerns for media provenance, misinformation, and digital forensics, as detection methods must reliably discriminate between real and generated content while remaining robust to the rapid evolution of new models.

Detecting AI-generated imagery is challenging for two intertwined reasons. First, generators increasingly match the \emph{visual} appearance of natural images, weakening cues based purely on texture or high-level semantics. Second, detector generalization is fragile: models trained on one set of generators often degrade on unseen ones, suggesting that many learned cues reflect generator- or dataset-specific biases rather than stable forensic evidence~\cite{ojha2024universal,cozzolino2024raisingbar}. Recent works have therefore explored complementary forensic clues such as frequency-domain artifacts~\cite{corvi2023intriguing,bammey2024synthbuster,karageorgiou2025spai} and inconsistencies in color statistics~\cite{uhlenbrock2024palette,jia2025secretcolor}. Color statistics are %a 
particularly appealing %forensic cue 
because real photographs are shaped by camera pipelines, including demosaicing, white balance, and tone mapping. In contrast, generative models are optimized mainly for perceptual fidelity in the RGB color space.

Motivated by these color-statistics discrepancies, we test the hypothesis that \emph{inter-channel dependencies} provide a compact and stable forensic cue.
In natural images, color channels co-vary due to the physics of image formation and camera pipelines; in contrast, generative models can match marginal color statistics while still deviating in how channels relate to each other, particularly in perceptually meaningful color spaces. Fig.~\ref{fig:cover_chroma} illustrates how $\mathrm{Lab}$ inter-channel correlation maps can surface traces that are visually subtle in RGB: when comparing a real photograph from RAISE-1k~\cite{dangnguyen2015raise} to a visually matched GPT-Image~1 rendering~\cite{openai_gptimage1}, the split correlation visualization reveals structured chromatic and texture discrepancies despite near-indistinguishable appearance in RGB.

We further probe whether such dependence is uniformly constrained by common perceptual objectives by analyzing how LPIPS~\cite{zhang2018lpips} responds to perturbations that selectively alter inter-channel coupling. We find that LPIPS exhibits non-uniform sensitivity to these changes, with responses strongly dependent on the color-space parameterization, indicating that cross-channel statistics are inconsistently constrained by perceptual similarity.

Motivated by this, we study the empirical distributions of pairwise inter-channel correlation features on semantically aligned real images from RAISE-1k~\cite{dangnguyen2015raise} and synthetic images from Synthbuster~\cite{bammey2024synthbuster}. The resulting distributional shifts are generator-dependent and representation-sensitive, with RGB and $\mathrm{Lab}$ providing clearer separation between real and generated sources than HSV or YUV in our setting.

Finally, we translate these observations into a lightweight detector, \textsc{Chroma}, which augments RGB inputs with inter-channel correlation maps and uses a fixed ResNet-50 backbone~\cite{he2016resnet}. Under a modest training budget, we show that $\mathrm{Lab}$ correlation cues provide complementary evidence beyond RGB appearance for real-vs-generated classification, and we study both single-generator training and a limited multi-generator supervision regime, where a small number of examples from additional generators is added to improve coverage of generator-specific modes. Across a standard benchmark protocol, correlation-augmented inputs improve robustness and yield competitive performance with recent detectors while preserving a simple architecture and training procedure.
\vspace*{-1em}

\section{Related Work}
\label{sec:related}
\vspace*{-1em}

Early work on synthetic image detection emerged in the GAN era and established foundational CNN-based detectors. Wang et al.~\cite{wang2020cnnspot} demonstrated reliable in-distribution detection for ProGAN-style synthesis, while Chai et al.~\cite{chai2020whatmakesfake} analyzed which cues transfer across datasets using patch-based models. Subsequent studies examined generalization limits~\cite{gragnaniello2021aregan}, improved robustness via multi-network training~\cite{mandelli2022orthogonal}, and proposed real-image statistics as a reference signal~\cite{liu2022detecting}. Collectively, these works clarified the strengths and limitations of GAN-era detectors and motivated the search for more stable forensic cues.

DeepFeatureX-SN~\cite{DeepFeatureX-SN} extends this paradigm by learning class-specific deep features for real, GAN-generated, and diffusion-generated images. The approach adopts a tripartite architecture in which three class-specific base models, implemented as Siamese CNNs trained via contrastive learning, extract discriminative features for real, GAN-generated, and diffusion-generated images. Features from the three branches are fed to a final CNN classifier.

Several works, on the other hand, seek to improve cross-generator robustness by decoupling representation learning from the detection task. Ojha et al.~\cite{ojha2024universal} demonstrate that end-to-end real/fake training promotes overfitting to generator-specific artifacts, resulting in poor generalization. Instead, they perform detection in the frozen feature space of a large vision–language model using nearest-neighbor search or shallow classifiers, achieving substantially improved transfer to unseen diffusion and autoregressive generators. Along similar lines, Cozzolino et al.~\cite{cozzolino2024raisingbar} demonstrate that reference-set decision rules applied to frozen CLIP embeddings yield strong cross-generator performance under controlled protocols. Lorenz et al.~\cite{lorenz2023detecting} further explore this idea using local intrinsic dimensionality estimated from fixed random network embeddings, achieving good in-distribution performance but limited transferability.

Several works emphasize that generalization depends not only on model architecture but critically on training data diversity and protocol design. In particular, Community Forensics~\cite{park2025communityforensics} demonstrates that detection performance improves monotonically with generator diversity, even when additional generators are architecturally similar diffusion models. Related studies confirm that detectors trained end-to-end can generalize to unseen generators to some extent, but performance degrades sharply under substantial architectural shifts~\cite{epstein2023online,gragnaniello2021aregan}.

Complementary work investigates what cues detection models actually exploit. Bird et al.~\cite{bird2024cifake} use Grad-CAM to analyze CNN-based detectors and show that predictions often rely on localized, low-salience artifacts rather than semantic content. Synthetic images tend to activate sparse regions, while real images exhibit more spatially distributed activations. These findings suggest that forensic cues often lie outside high-level semantics, motivating the study of low-level, interpretable image statistics.

Beyond pure image backbones, Huang et al.~\cite{huang2025sida} leverage large multimodal models for joint detection, localization, and explanation. Their SIDA framework is formulated as a multi-task system that, given an image, produces an authenticity prediction, a localization map of manipulated regions, and a textual rationale for its decision. This approach shows one way in which large pre-trained multimodal models can be adapted to combine detection, localization, and explanation, while still operating primarily on generic high-level image representations rather than explicitly designed forensic cues.

% Another major line of research exploits frequency-domain artifacts left by generative models. Corvi et al.~\cite{corvi2023intriguing} show that both GANs and diffusion models exhibit characteristic spectral signatures,  arising from generator architectures and inherited dataset biases. Building on these observations, several methods directly leverage spectral traces for detection. Synthbuster~\cite{bammey2024synthbuster} targets diffusion outputs by analyzing Fourier-domain peaks of high-pass residuals while explicitly accounting for compression artifacts. MaskSim~\cite{masksim} jointly learns discriminative frequency masks and reference spectral patterns, yielding interpretable detection cues. SPAI~\cite{karageorgiou2025spai} models the spectral distribution of real images in a self-supervised manner and treats synthetic images as out-of-distribution samples, while FIRE~\cite{chu2025fire} exploits reconstruction errors in selectively filtered frequency bands. Together, these works demonstrate that generator-induced spectral regularities persist despite high visual realism, though they often rely on carefully engineered filtering or reconstruction strategies.

Another major line of research exploits frequency-domain artifacts left by generative models.Corvi et al.~\cite{corvi2023intriguing} show that GANs and diffusion models exhibit characteristic spectral signatures driven by architecture and training-data biases. Building on this, Synthbuster~\cite{bammey2024synthbuster} detects diffusion images via Fourier peaks of high-pass residuals while accounting for compression, and MaskSim~\cite{masksim} learns discriminative frequency masks and reference spectra for interpretable cues. SPAI~\cite{karageorgiou2025spai} instead models real-image spectral statistics self-supervisedly and flags synthetics as out-of-distribution, while FIRE~\cite{chu2025fire} leverages reconstruction errors in selected frequency bands. Collectively, these results indicate that spectral regularities can persist despite high visual realism, albeit often requiring careful filtering or reconstruction.

Related low-level formulations include gradient/residual representations that suppress semantic content while amplifying synthesis traces. LGrad~\cite{tan2023lgrad} learns on image gradients as a generalized artifact representation, and Tan et al.~\cite{tan2024upsampling} revisit upsampling operations to improve generalization in CNN-based deepfake detection. Orthogonally, DIRE~\cite{wang2023dire} performs diffusion-based reconstruction and uses reconstruction behavior as a detection cue, providing a generative-prior alternative to purely discriminative features.

Discrepancies in color statistics between real and synthetic images were first reported for GANs by Li et al.~\cite{li2020identification}, who show that inconsistencies become more pronounced in chrominance representations than in RGB. Subsequent work extends these observations to diffusion models. Uhlenbrock et al.~\cite{uhlenbrock2024palette} conduct a systematic analysis of color distributions across multiple color spaces, arguing that perceptual training objectives are more sensitive to luminance than to chrominance variations. By designing simple relational color transforms, they expose visually interpretable artifacts and build a lightweight detector using hand-crafted features. Jia et al.~\cite{jia2025secretcolor} propose a complementary approach that measures color-distribution uniformity through repeated noise injection and quantization, showing that real images exhibit more stable color statistics than synthetic ones.

In contrast to prior color-based methods that primarily model marginal color distributions or rely on hand-crafted color-space transformations, our work targets \emph{inter-channel dependencies} explicitly, as a compact and interpretable forensic cue, and systematically studies their behavior across color-space parameterizations.

\vspace*{-1em}
\section{Inter-Channel Correlations as Forensic Cues}
\label{sec:inter_channel_correlation}

Within natural images, inter-channel dependencies provide a compact description of how color components co-vary. Unlike spatial texture cues, these statistics are directly tied to the structure induced by camera pipelines and color formation processes, and can be inspected across different color parameterizations. In this section, we study inter-channel correlations as a forensic trace for AI-generated image detection. We first analyze how a standard perceptual similarity metric reacts to controlled perturbations that modify channel dependence, and later examine how correlation distributions differ between real and generated images across multiple color spaces.
\vspace*{-1em}

\subsection{LPIPS Sensitivity to Inter-Channel Dependence}

Perceptual similarity metrics such as LPIPS~\cite{zhang2018lpips} are deeply intertwined with modern generative modeling. In addition to being reported at evaluation time, perceptual feature 
losses are routinely used when adapting or fine-tuning diffusion pipelines for inverse problems and image editing (e.g., reconstruction, inpainting, or image-to-image tasks), where matching human-perceived similarity cannot be captured by pixel-wise criteria alone. In contrast, LPIPS is computed from the deep features of RGB images, and its sensitivity to chromatic structure depends on how a perturbation manifests when mapped back to RGB.

From this observation, and following the color-relation perspective of~\cite{uhlenbrock2024palette}, we investigate whether LPIPS consistently reflects changes in inter-channel dependence across color spaces. We design controlled perturbations that either preserve channel coupling or explicitly decorrelate channels in RGB, Lab, HSV, and YUV, and measure the induced LPIPS distances. If LPIPS is uneven in this respect, then perceptual objectives may fail to enforce correlation structure uniformly, motivating the use of correlation-based descriptors for detection.

To analyze LPIPS sensitivity to inter-channel dependence, we apply controlled, low-amplitude perturbations to RGB images and measure the induced perceptual distance. 
Let $x$ denote a natural RGB image. We construct two perturbed versions of $x$ by adding noise on a subset of pixels comprising a fraction $\rho$ of the image. 
In the first case, we add \emph{channel-correlated} noise by sampling a single noise field $n$ with i.i.d.\ entries and adding it identically to the three channels:
\[
x^{\text{corr}}_c = x_c + n, \quad c \in \{R,G,B\}.
\]
In the second case, we add \emph{decorrelated} noise by sampling three independent noise fields $n_R, n_G, n_B$ with the same marginal distribution and adding them channel-wise:
\[
x^{\text{decorr}}_c = x_c + n_c, \quad c \in \{R,G,B\}.
\]
In both settings, we use zero-mean Gaussian noise with standard deviation $\sigma = 0.005$ in normalized $[0,1]$ RGB intensities, yielding visually imperceptible perturbations, and we vary $\rho$ to control the fraction of affected pixels. 
We then compute the LPIPS distance between each perturbed image and its corresponding original in RGB space, obtaining LPIPS responses as a function of $\rho$ under correlated and decorrelated noise.

This procedure is repeated over the full RAISE-1k dataset~\cite{dangnguyen2015raise}. For each $\rho$, we report the mean LPIPS across all images, yielding dataset-level response curves.

\begin{figure}[t]
    \centering
    \includegraphics[width=0.7\linewidth]{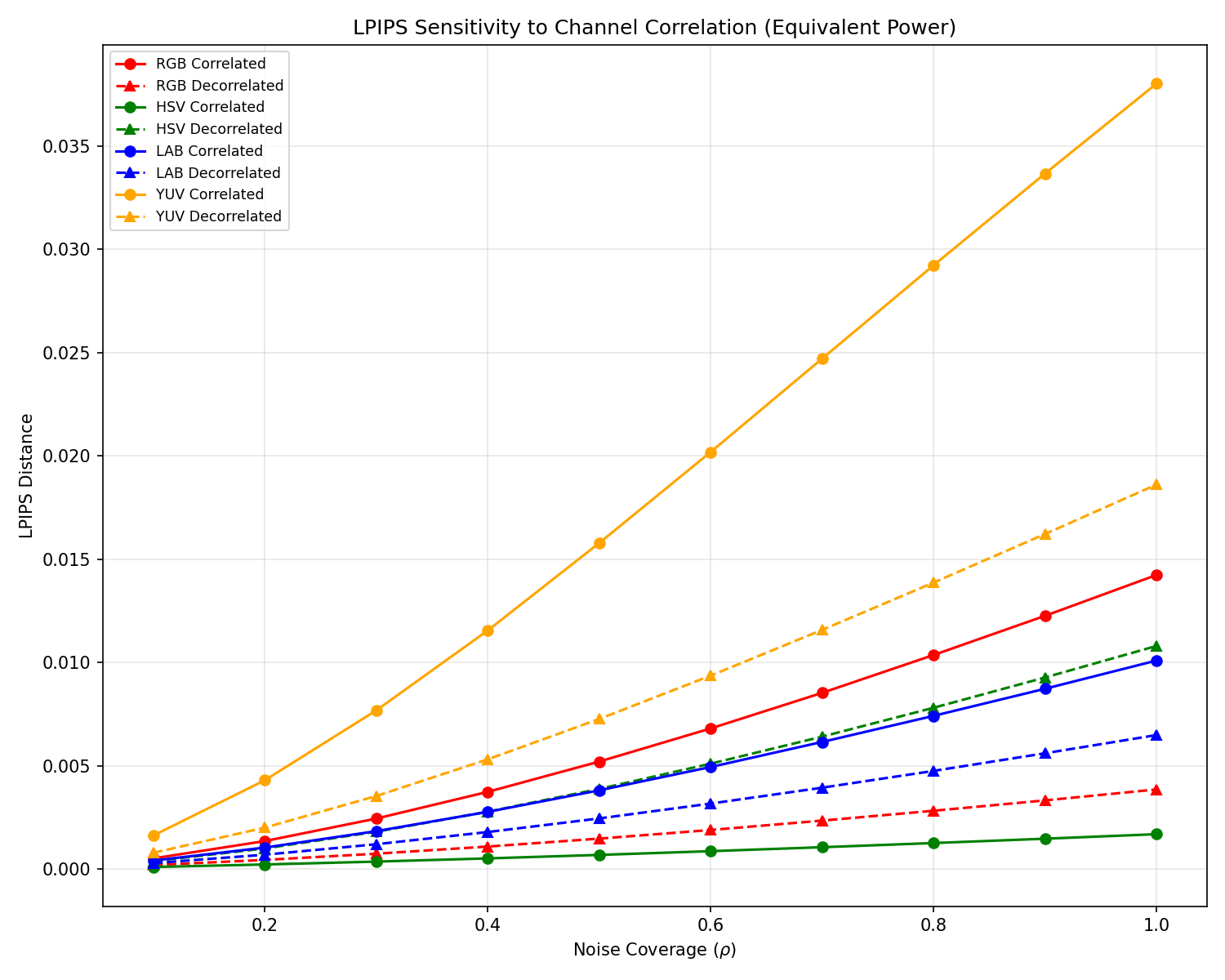}
    \vspace*{-1em}
    \caption{RAISE-1k-averaged LPIPS distance between an original RGB image and its perturbed versions as a function of noise coverage~$\rho$ for different color spaces (RGB, Lab, HSV, YUV) under correlated (solid) and decorrelated (dashed) channel noise with matched perturbation power.}
    \label{fig:lpips_color_spaces}
    \vspace*{-1.2em}
\end{figure}

We extend the same analysis to alternative color spaces, including Lab, YUV, and HSV. Given an RGB image, we convert it to a target color space, apply correlated and decorrelated noise to its channels as above, and transform the perturbed result back to RGB before computing LPIPS. To ensure comparability across color spaces, the noise variance in each space is adjusted so that, after conversion back to RGB, the induced perturbations have the same effective magnitude $\sigma = 0.005$ as the RGB baseline. The resulting dataset-averaged curves are summarized in Fig.~\ref{fig:lpips_color_spaces}.

Fig.~\ref{fig:lpips_color_spaces} reveals that LPIPS responds very differently to perturbations that are matched in \emph{effective RGB power} but injected in different color spaces. First, for a fixed coverage~$\rho$, the LPIPS magnitude varies substantially across color spaces: YUV consistently yields the largest distances, whereas RGB, Lab, and HSV produce smaller responses, with their relative ordering depending on whether the injected noise is correlated or decorrelated. Since the perturbations are power-matched after reconstruction to RGB, this spread cannot be explained by the magnitude of noise, but rather by how LPIPS weights the resulting distortions once they are induced through different channel parameterizations. In other words, LPIPS is not invariant to reparameterizations that redistribute variation across luminance and chrominance components.

Second, and more importantly for our motivation, the metric is \emph{inconsistent} in how it treats changes in channel dependence. The gap between correlated and decorrelated noise is pronounced in some spaces (most clearly YUV, and to a lesser extent RGB and HSV), but it nearly collapses in Lab. This means that two perturbations that explicitly differ in whether they preserve or disrupt inter-channel structure can be assigned almost the same perceptual distance depending on the coordinate system in which the perturbation is expressed.

These behaviors suggest that LPIPS can under-emphasize certain chromatic \emph{joint-statistics} changes—particularly those expressed in perceptually meaningful coordinates, such as Lab, while being more sensitive to others, such as YUV. Consequently, when generators are optimized using such perceptual objectives, matching inter-channel correlation structure is not uniformly encouraged across color spaces. This motivates studying inter-channel correlation statistics directly: if these dependencies are only weakly constrained by perceptual losses, they may persist as systematic, generator-dependent discrepancies that can be exploited as forensic evidence for discriminating between real and generated images.
\vspace*{-1em}

\subsection{Cross-Generator Distributions of Inter-Channel Correlations}
\label{subsec:corr_distributions}

% Given that LPIPS reacts unevenly to perturbations that alter inter-channel dependence across color spaces, we next examine whether these correlation statistics themselves exhibit systematic, source-dependent differences that can be exploited for detection.
% For each image, we transform it to a given color space (RGB, HSV, Lab, YUV), compute the three pairwise inter-channel correlation maps, and aggregate the resulting values over the dataset to obtain one empirical distribution per source (real vs.\ each generator). Fig.~\ref{fig:correlation_distributions} shows \emph{log-density} estimates for each channel pair, comparing real images from RAISE-1k~\cite{dangnguyen2015raise} to synthetic images generated by DALL{\text -}E, GLIDE, Midjourney v5, Firefly, and Stable Diffusion from Synthbuster~\cite{bammey2024synthbuster}. Importantly, Synthbuster was built to match the semantic content of its synthetic images to the real-image reference set, reducing scene-dependent variation and helping to attribute distribution shifts to the generative process rather than to differences in depicted content.

Given that LPIPS reacts unevenly to perturbations that alter inter-channel dependence across color spaces, we next examine whether these correlation statistics themselves exhibit systematic, source-dependent differences that can be exploited for detection. 
For each image and each color space, we construct local inter-channel correlation maps by computing the Pearson correlation coefficient over sliding $k\times k$ windows, so that each map value captures channel dependence within a local neighborhood. In our experiments, we set 
$k=7$.
%For each image and each color space, we compute \emph{local} inter-channel correlation maps using a sliding-window Pearson correlation. 
%Concretely, for a channel pair $(a,b)$ we estimate, at each pixel location, the local means and second-order moments within a $k\times k$ neighborhood (implemented via average pooling) with $k=7$, and form the normalized covariance
%\[
%\mathrm{Corr}_{a,b}(p) \;=\; \frac{\mathrm{Cov}_{a,b}(p)}{\sqrt{\mathrm{Var}_{a}(p)}\sqrt{\mathrm{Var}_{b}(p)}+\varepsilon}\,,
%\]
%yielding a correlation value in $[-1,1]$ per pixel.

\begin{figure*}[ht!]
    \centering
    \captionsetup[subfigure]{font=scriptsize}

    % -------------------- RGB row --------------------
    \begin{subfigure}[t]{\textwidth}
        \centering
        \includegraphics[width=0.325\linewidth]{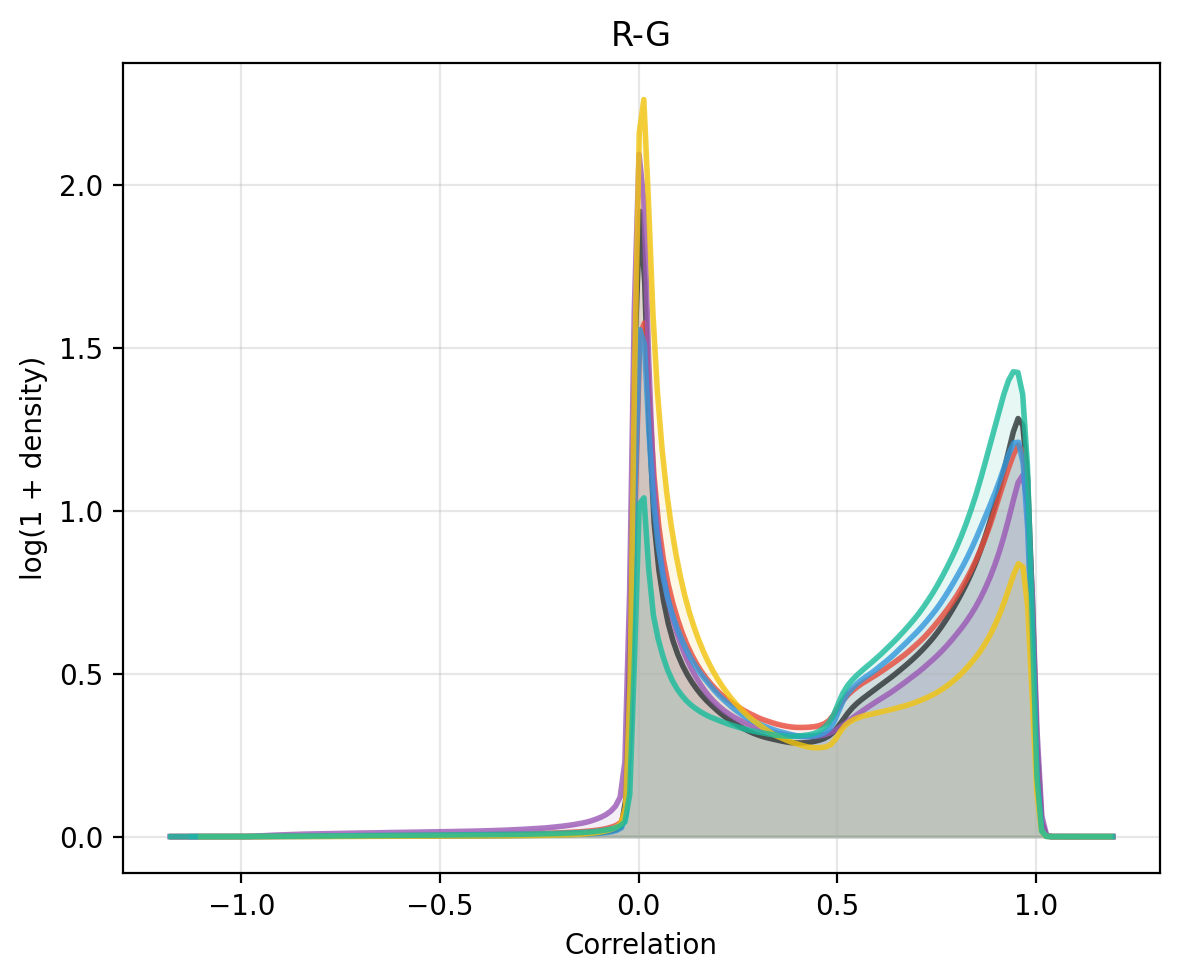}\hfill
        \includegraphics[width=0.325\linewidth]{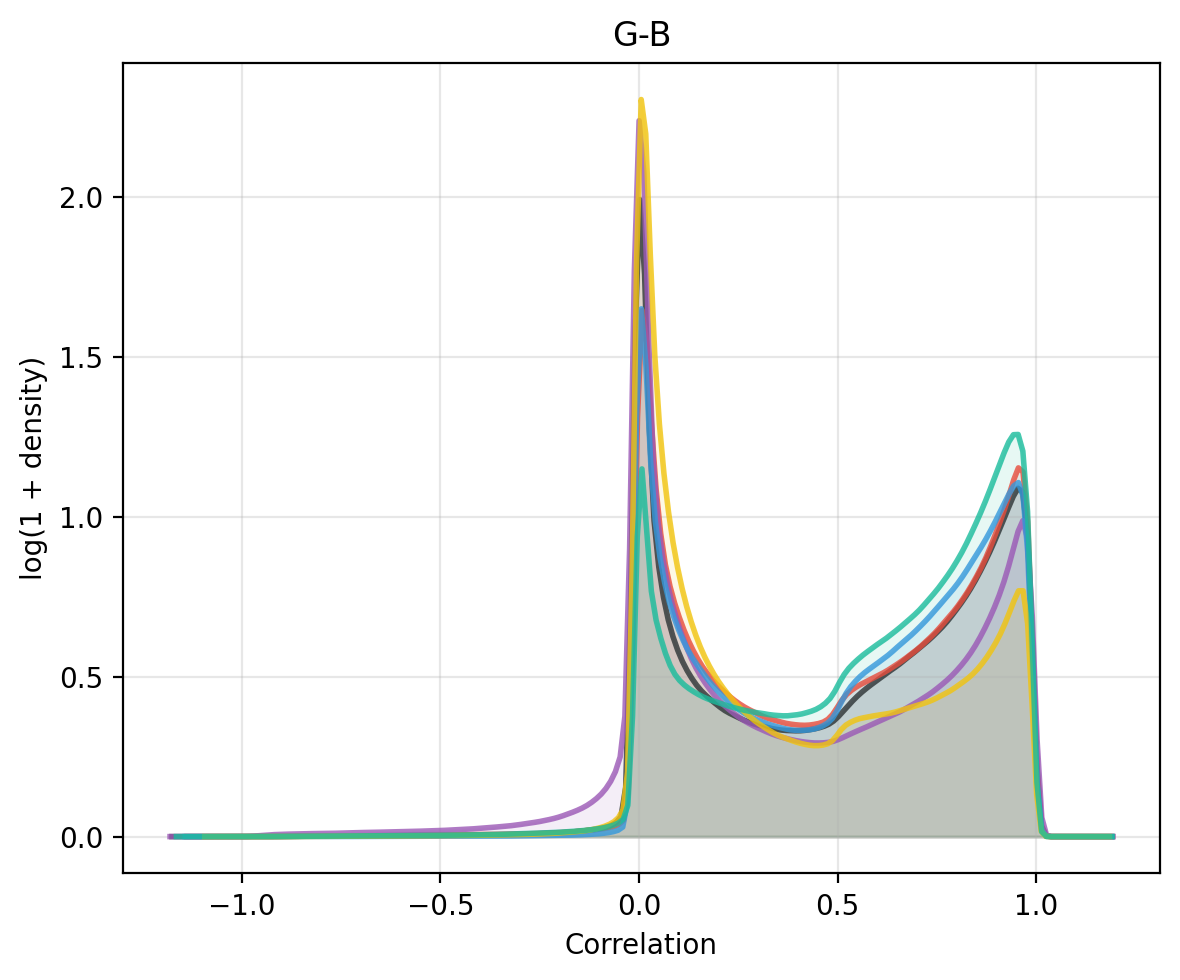}\hfill
        \includegraphics[width=0.325\linewidth]{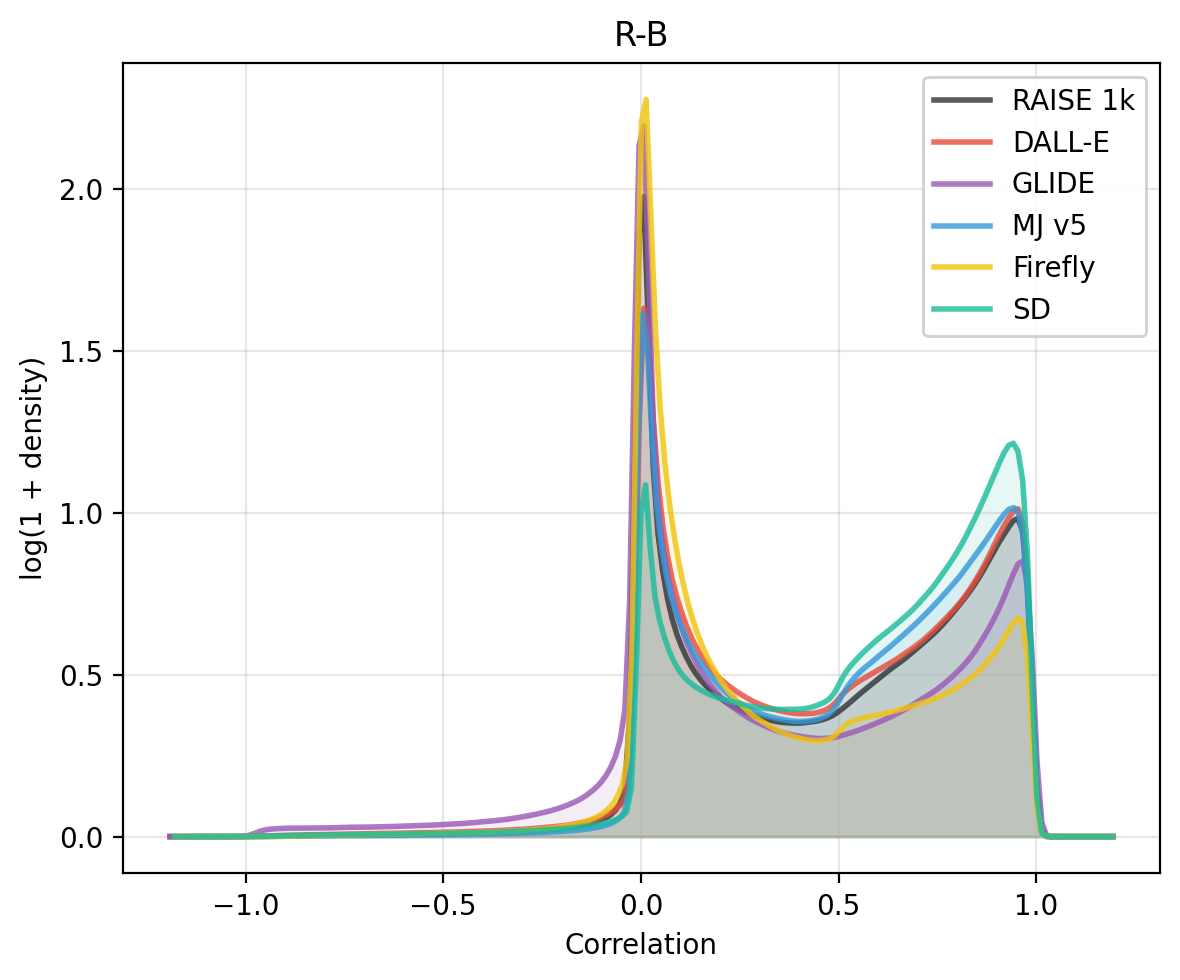}
        \vspace*{-0.6em}
        % \caption{RGB Inter-Channel Correlations}
    \end{subfigure}

    \vspace{0.6em}

    % -------------------- HSV row --------------------
    \begin{subfigure}[t]{\textwidth}
        \centering
        \includegraphics[width=0.325\linewidth]{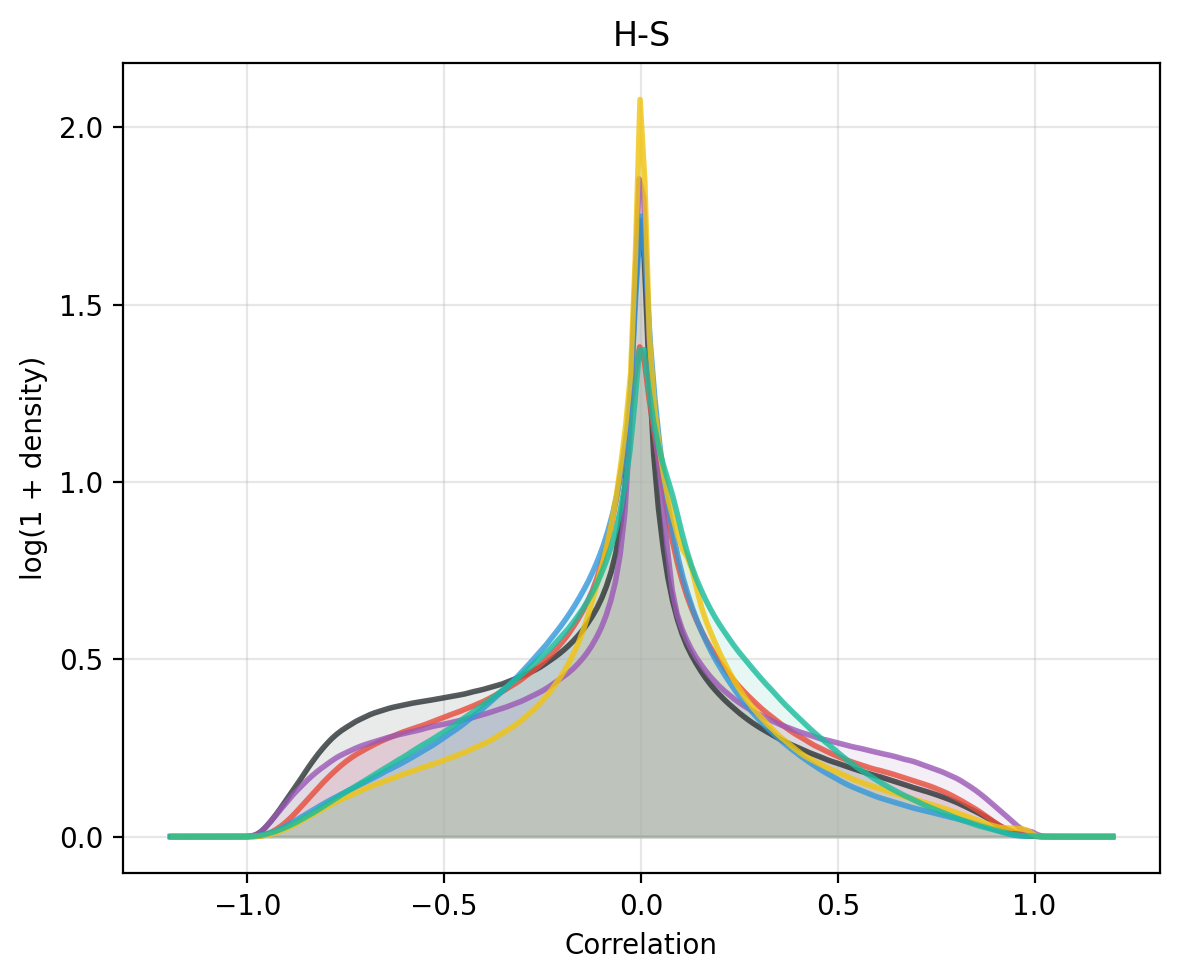}\hfill
        \includegraphics[width=0.325\linewidth]{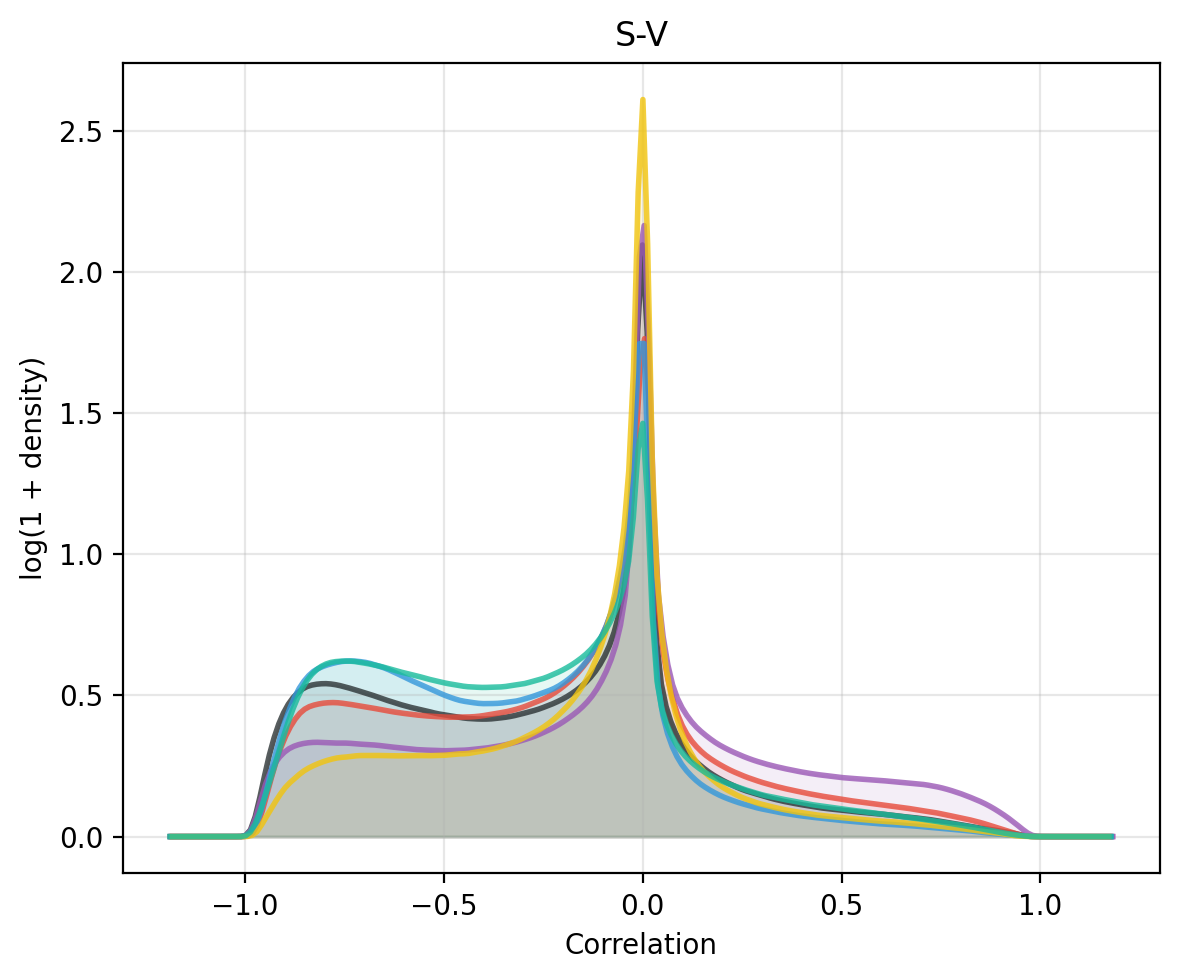}\hfill
        \includegraphics[width=0.325\linewidth]{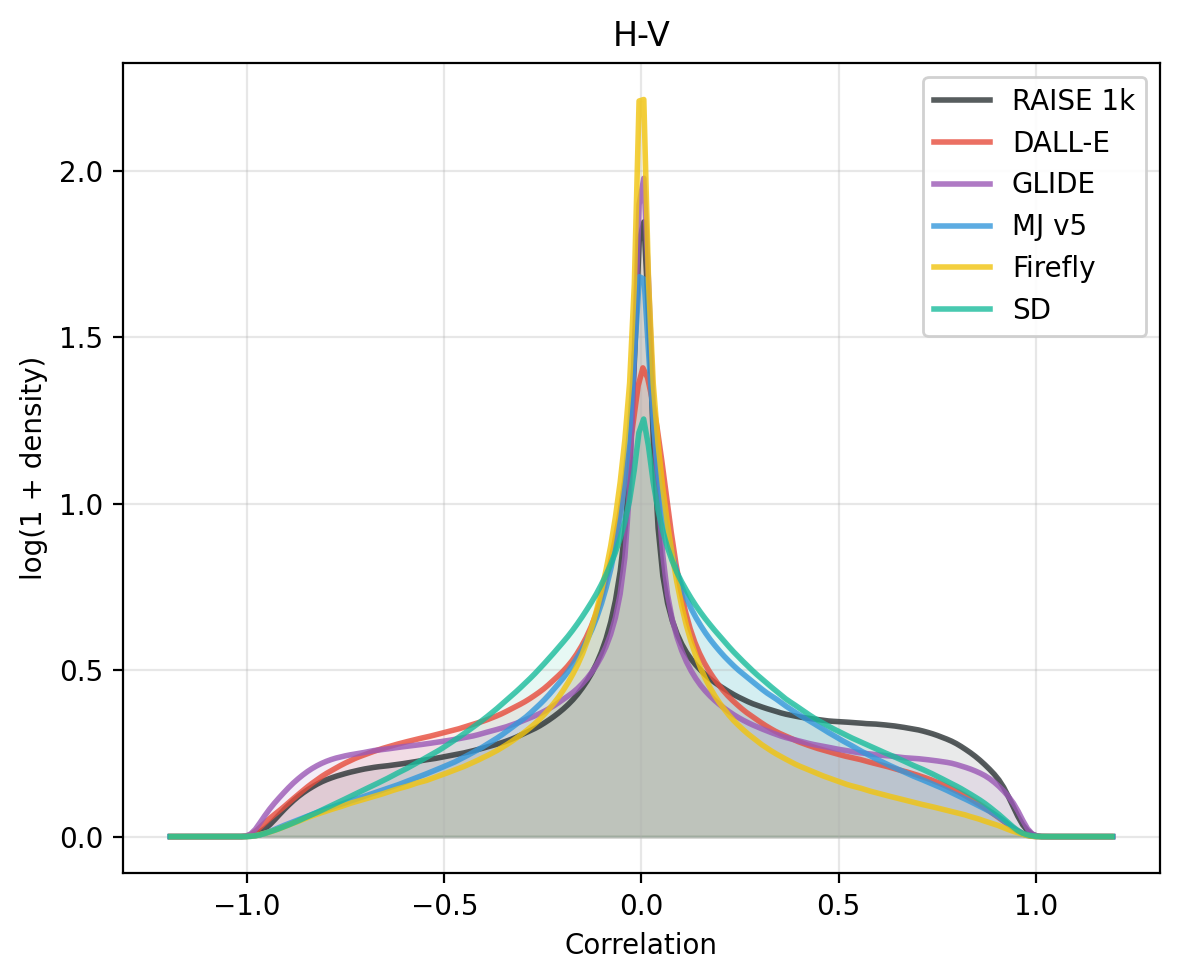}
        \vspace*{-0.6em}
        % \caption{HSV Inter-Channel Correlations}
    \end{subfigure}

    \vspace{0.6em}

    % -------------------- Lab row --------------------
    \begin{subfigure}[t]{\textwidth}
        \centering
        \includegraphics[width=0.325\linewidth]{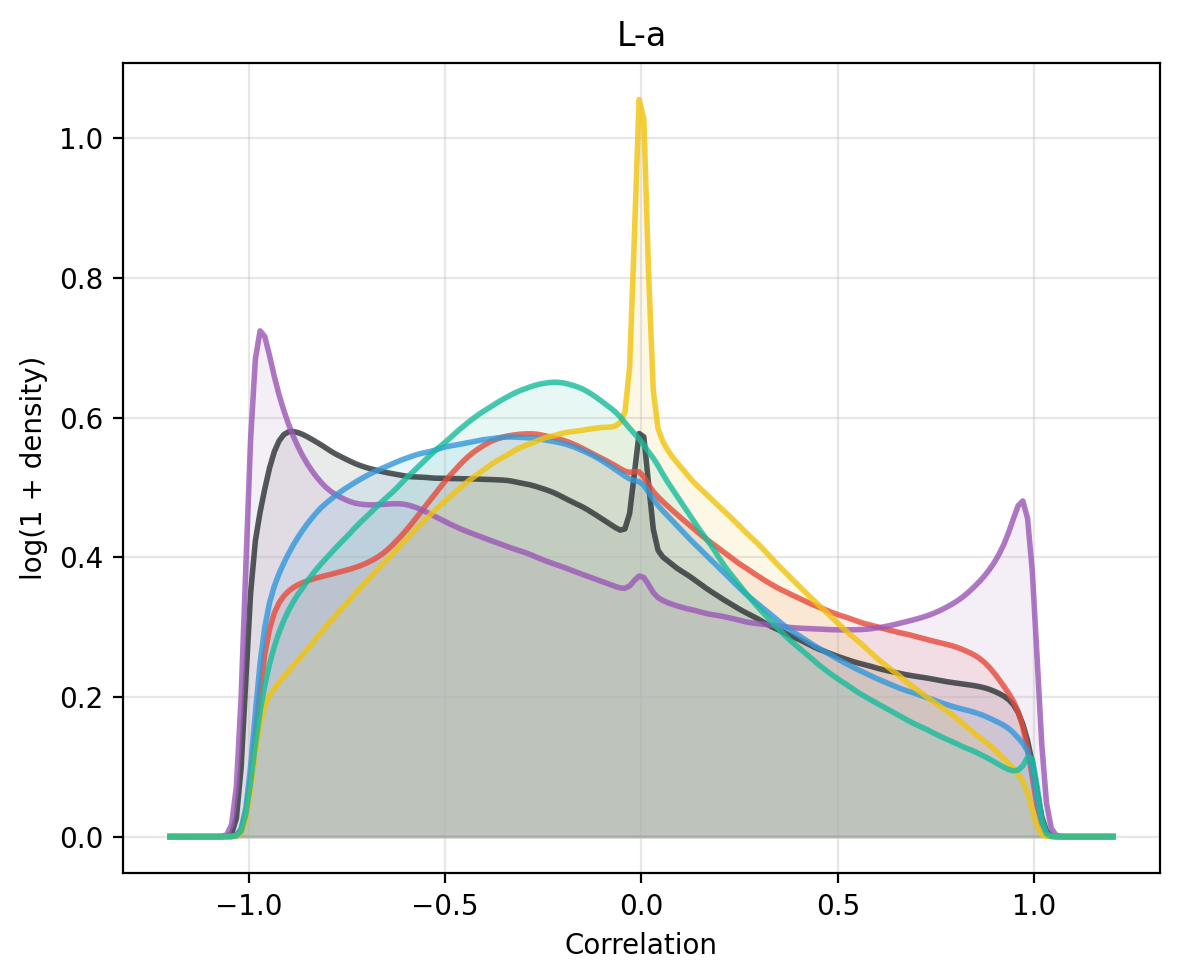}\hfill
        \includegraphics[width=0.325\linewidth]{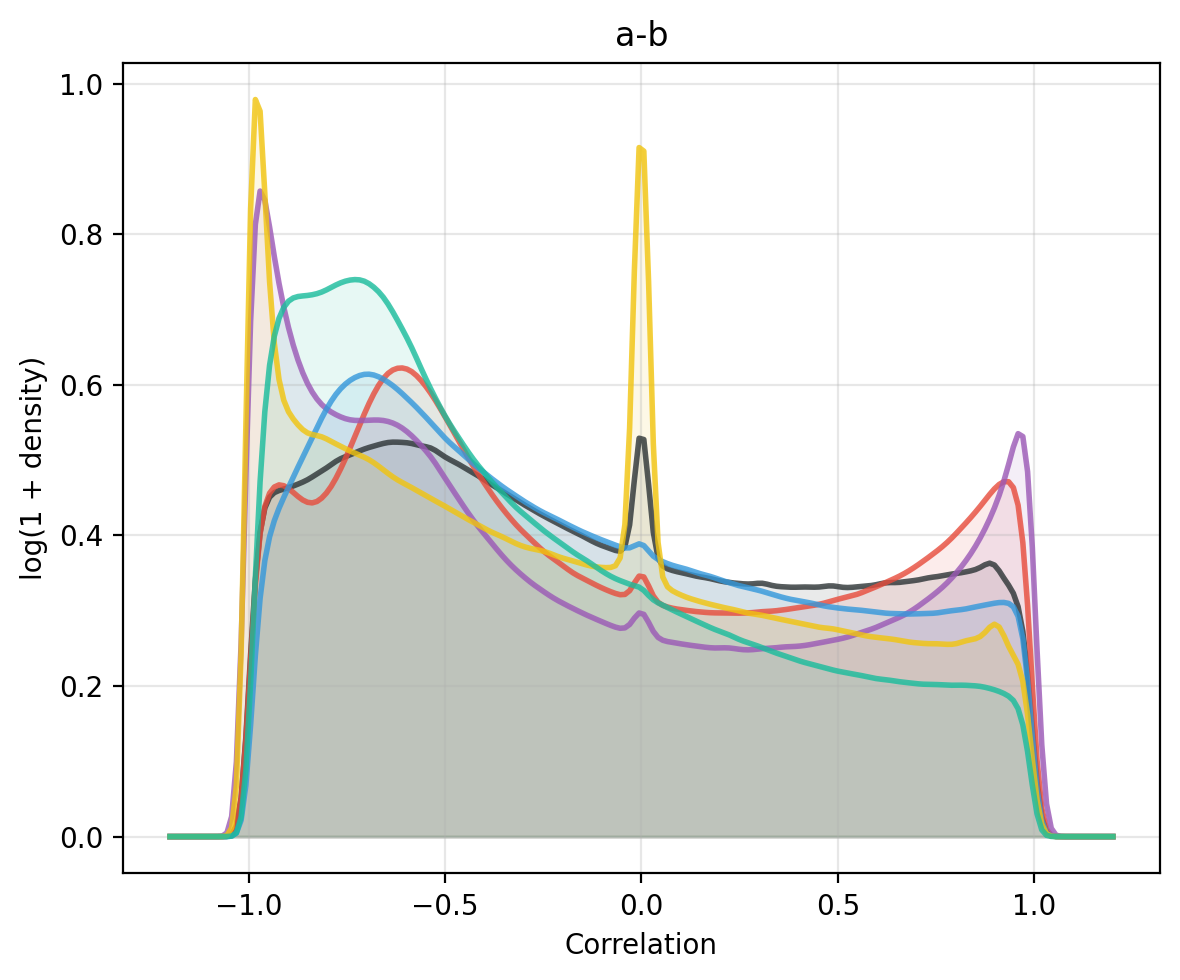}\hfill
        \includegraphics[width=0.325\linewidth]{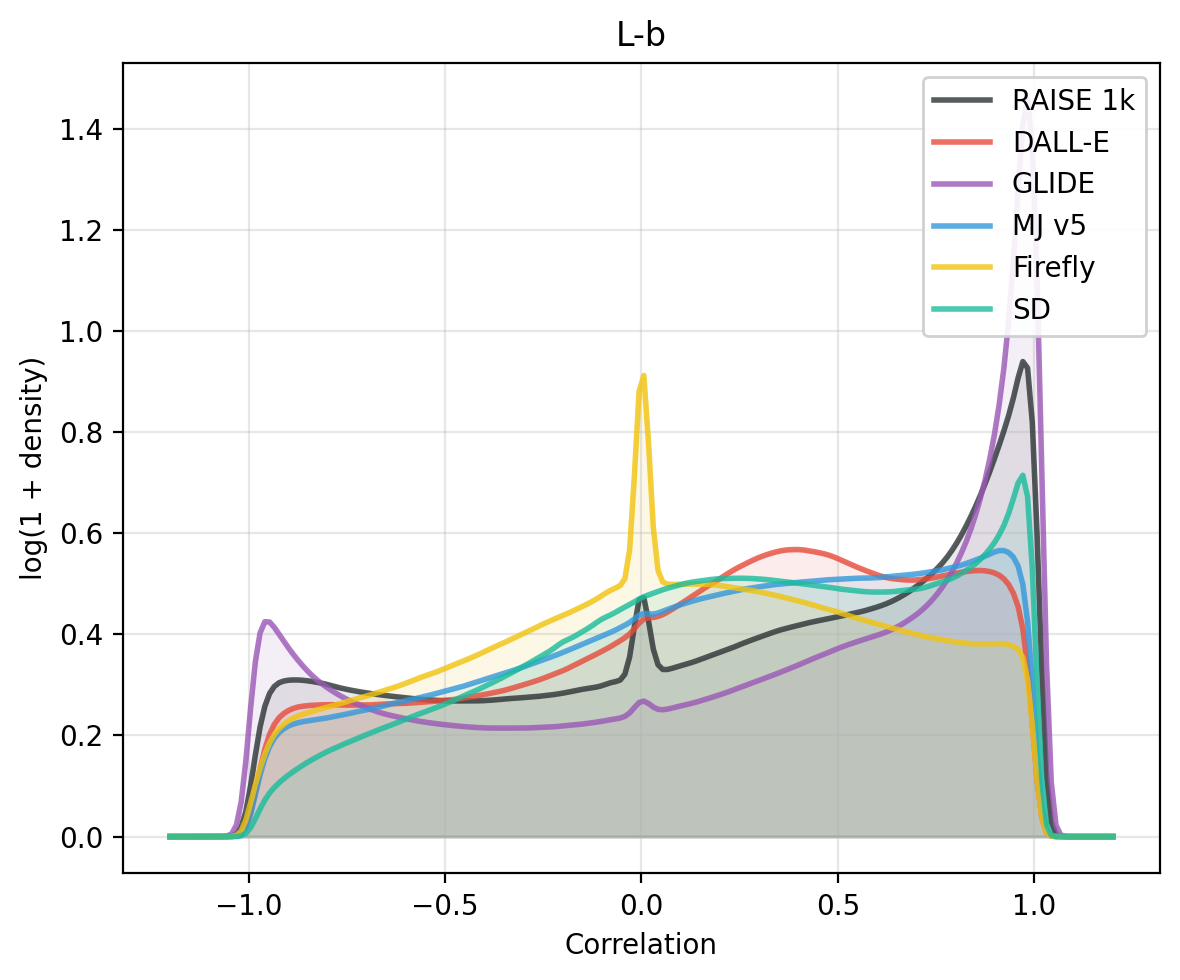}
        \vspace*{-0.6em}
        % \caption{Lab Inter-Channel Correlations}
    \end{subfigure}

    \vspace{0.6em}

    % -------------------- YUV row --------------------
    \begin{subfigure}[t]{\textwidth}
        \centering
        \includegraphics[width=0.325\linewidth]{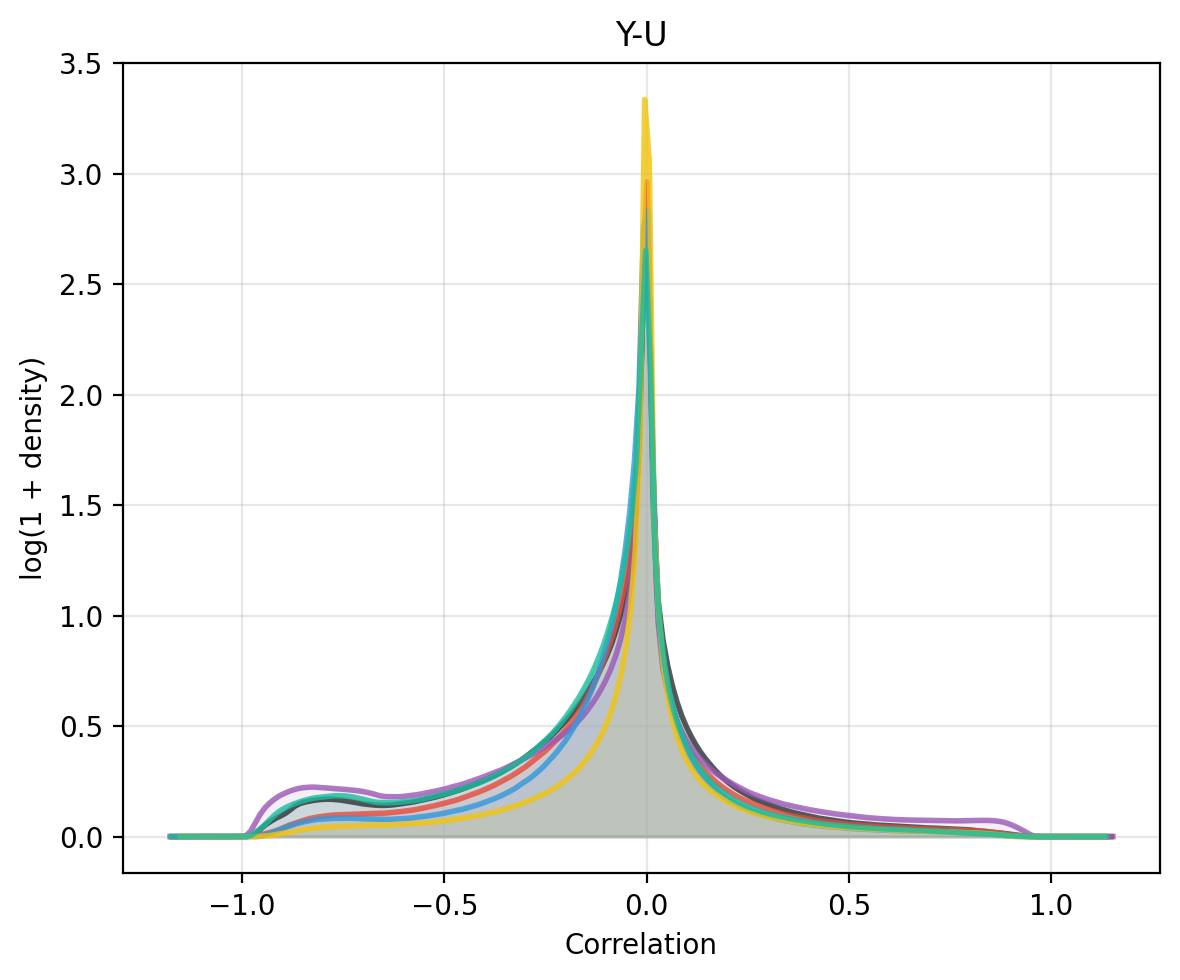}\hfill
        \includegraphics[width=0.325\linewidth]{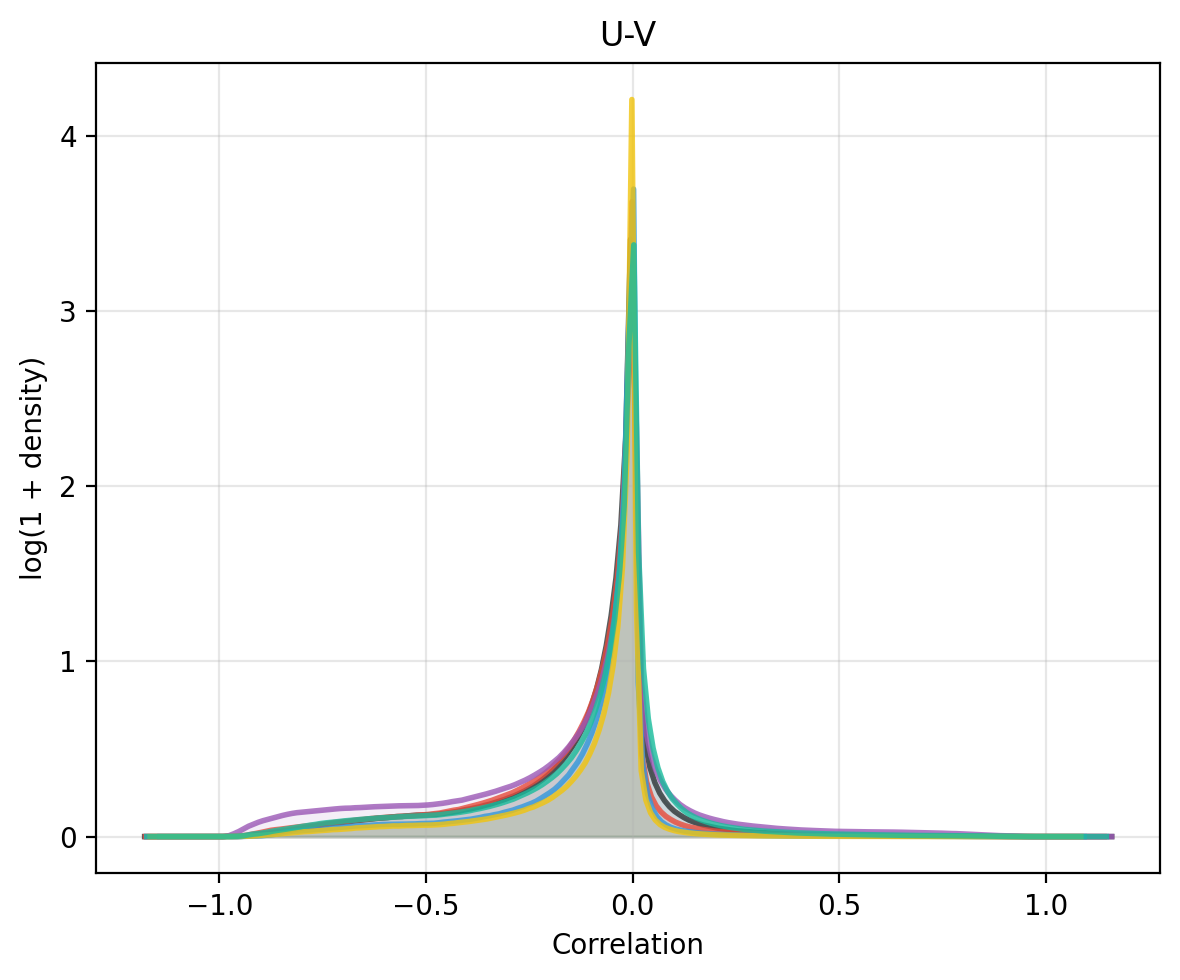}\hfill
        \includegraphics[width=0.325\linewidth]{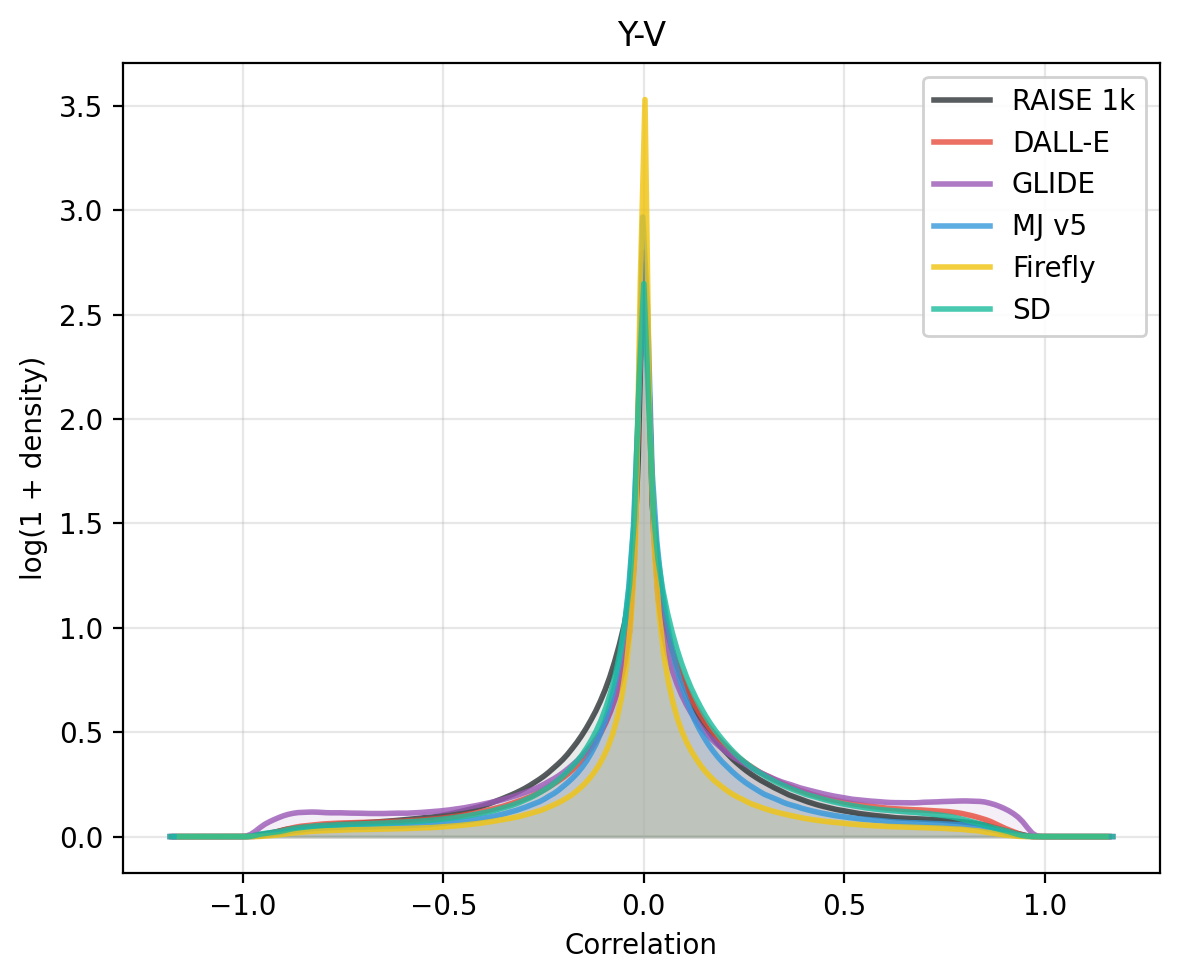}
        \vspace*{-0.6em}
        % \caption{YUV Inter-Channel Correlations}
    \end{subfigure}

    \caption{Log-density estimates of pairwise inter-channel correlation features across color spaces (RGB, HSV, Lab, YUV).
    Each row corresponds to one color space and shows the three channel-pair correlations.
    We report real images from RAISE-1k~\cite{dangnguyen2015raise} and synthetic images from multiple generators in Synthbuster~\cite{bammey2024synthbuster}.}
    \label{fig:correlation_distributions}
    \vspace*{-2em}
\end{figure*}

For each color space, %(RGB, HSV, Lab, YUV), 
we compute three such maps corresponding to the three channel pairs, and then aggregate correlation values over all pixels and images in a dataset to obtain one empirical distribution per source.
% (real vs.\ each generator).
Fig.~\ref{fig:correlation_distributions} shows the \emph{log-density} estimates for each channel pair, comparing real images from RAISE-1k~\cite{dangnguyen2015raise} to synthetic images generated by DALL{\text -}E, GLIDE, Midjourney v5, Firefly, and Stable Diffusion from Synthbuster~\cite{bammey2024synthbuster}. 
Importantly, Synthbuster was designed to align the semantic content of its synthetic images with the real-image reference set, thereby reducing scene-dependent variation and enabling the attribution of distribution shifts to the generative process rather than to differences in depicted content.

Several patterns are visible in Fig.~\ref{fig:correlation_distributions}. For RGB and HSV, the curves share a dominant mode and largely overlap, with generator-dependent differences appearing mainly in the tails and in the relative mass around the peak. In Lab, the separation is more pronounced: correlations involving the luminance channel (L) exhibit clearer generator-specific shifts and shape changes, indicating that Lab correlations capture deviations that are less apparent in RGB/HSV. In contrast, YUV correlations are tightly concentrated around a narrow region for most sources, leading to smaller visible shifts across generators.

To summarize these differences quantitatively, Fig.~\ref{fig:corr_wasserstein_matrices} reports pairwise Wasserstein distances between the correlation-feature distributions for each color space; for a given color space, the reported value is the average of the three pairwise inter-channel Wasserstein distances. The matrices confirm that separability depends strongly on the representation: RGB and Lab show consistently larger off-diagonal distances than HSV and YUV, and they also exhibit a clearer generator-specific structure. For example, in RGB the largest gaps arise for pairs involving Firefly and Stable Diffusion (e.g., Firefly--SD), while Lab also yields substantial separation for several generator pairs (e.g., GLIDE vs.\ Firefly/SD), reflecting the broader and more shifted Lab log-density curves. By comparison, YUV distances remain systematically smaller across the board, consistent with the tighter concentration of its correlation distributions.

\newcommand{\matH}{0.16\textheight}
\newcommand{\matimgH}[1]{\includegraphics[height=\matH,keepaspectratio]{#1}}
\newcommand{\subcapshift}{1.2em}

\begin{figure*}[t]
    \centering
    \captionsetup[subfigure]{font=scriptsize}
    \setlength{\tabcolsep}{4pt}
    \renewcommand{\arraystretch}{1.0}

    \begin{tabular}{@{}c c c c@{}}
        \begin{subfigure}[t]{0.22\textwidth}
            \centering
            \matimgH{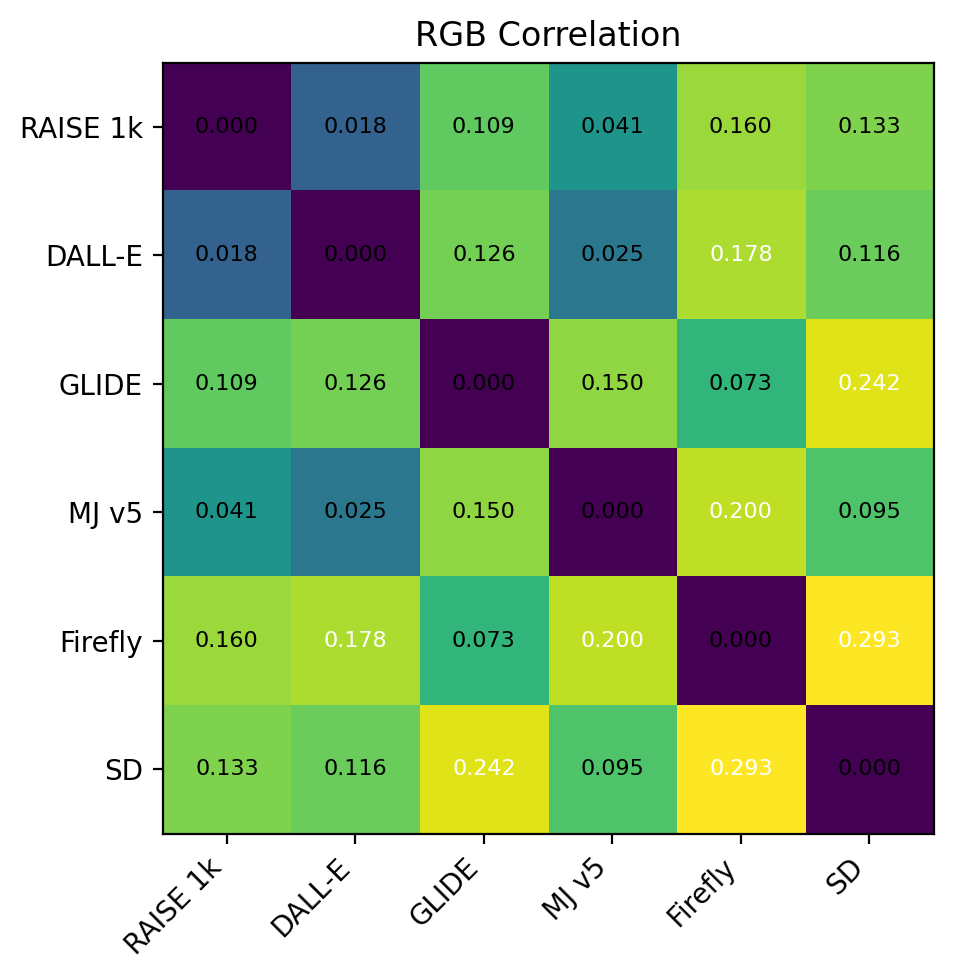}
            \vspace*{-2em}
            \caption{\hspace*{\subcapshift}RGB}
        \end{subfigure}
        &
        \begin{subfigure}[t]{0.22\textwidth}
            \centering
            \matimgH{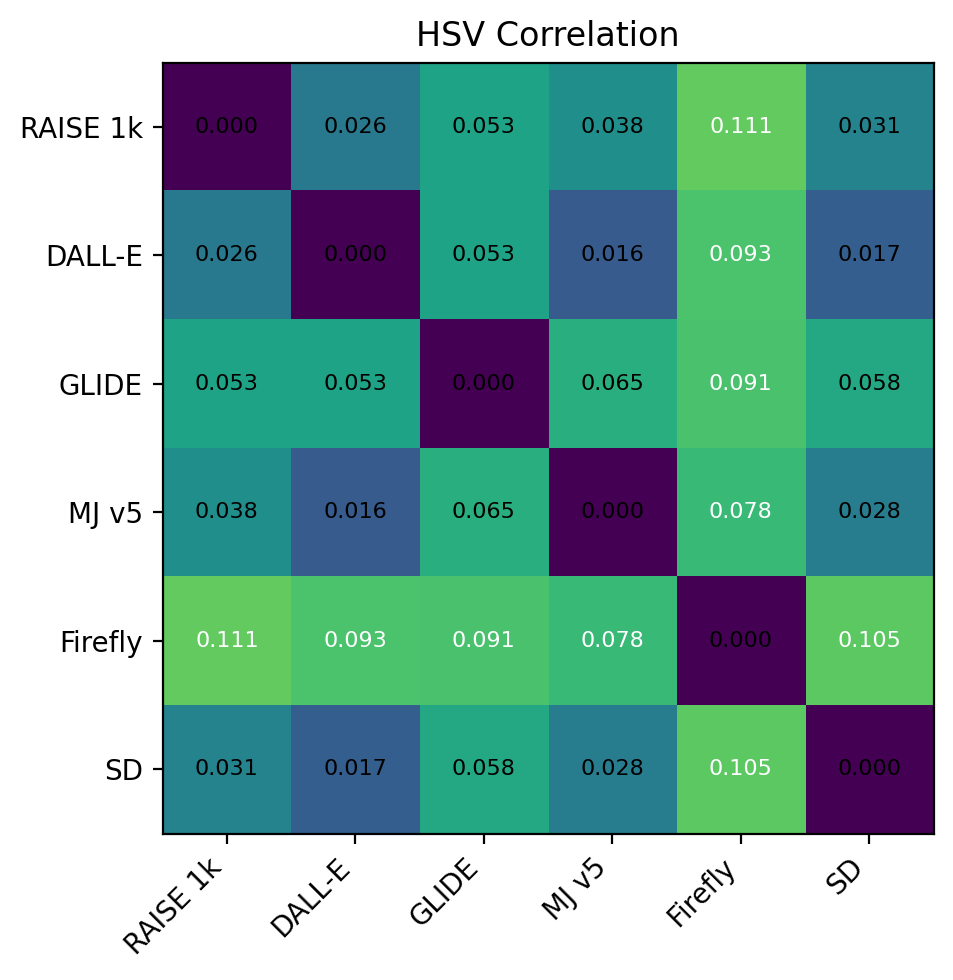}
            \vspace*{-2em}
            \caption{\hspace*{\subcapshift}HSV}
        \end{subfigure}
        &
        \begin{subfigure}[t]{0.22\textwidth}
            \centering
            \matimgH{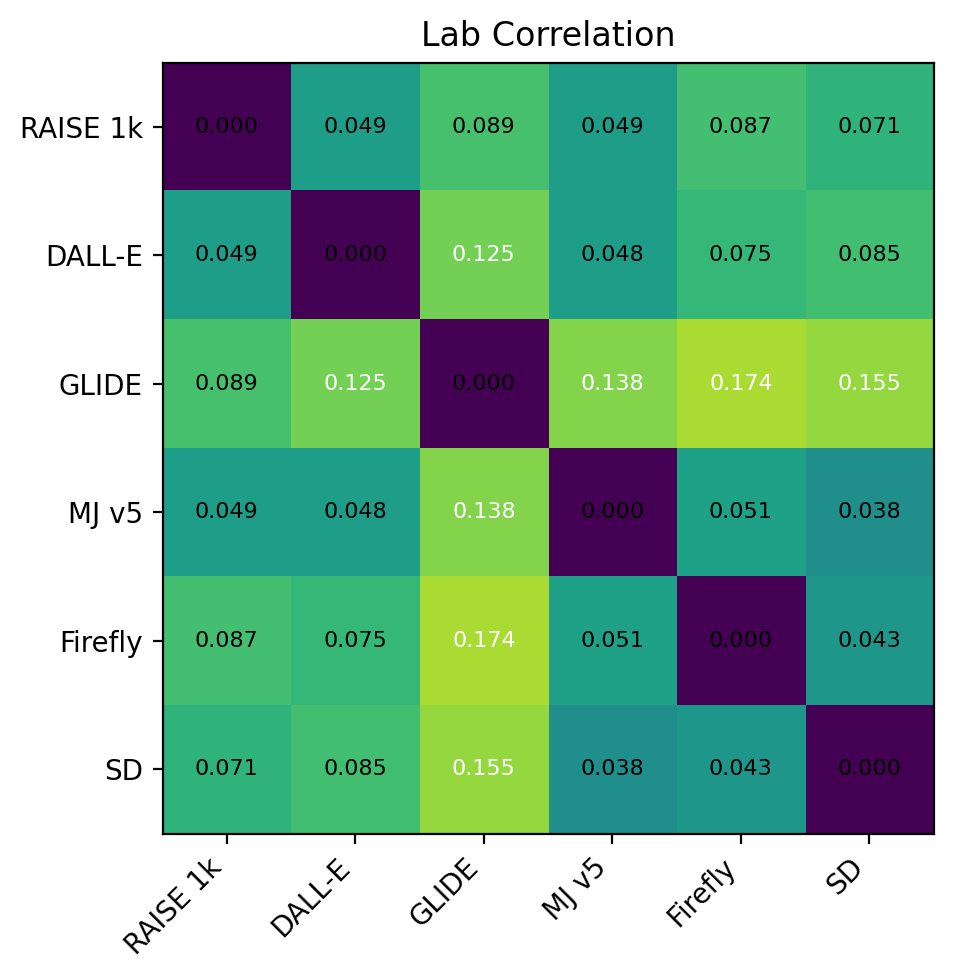}
            \vspace*{-2em}
            \caption{\hspace*{\subcapshift}Lab}
        \end{subfigure}
        &
        \begin{subfigure}[t]{0.30\textwidth} % wider for colorbar
            \centering
            \matimgH{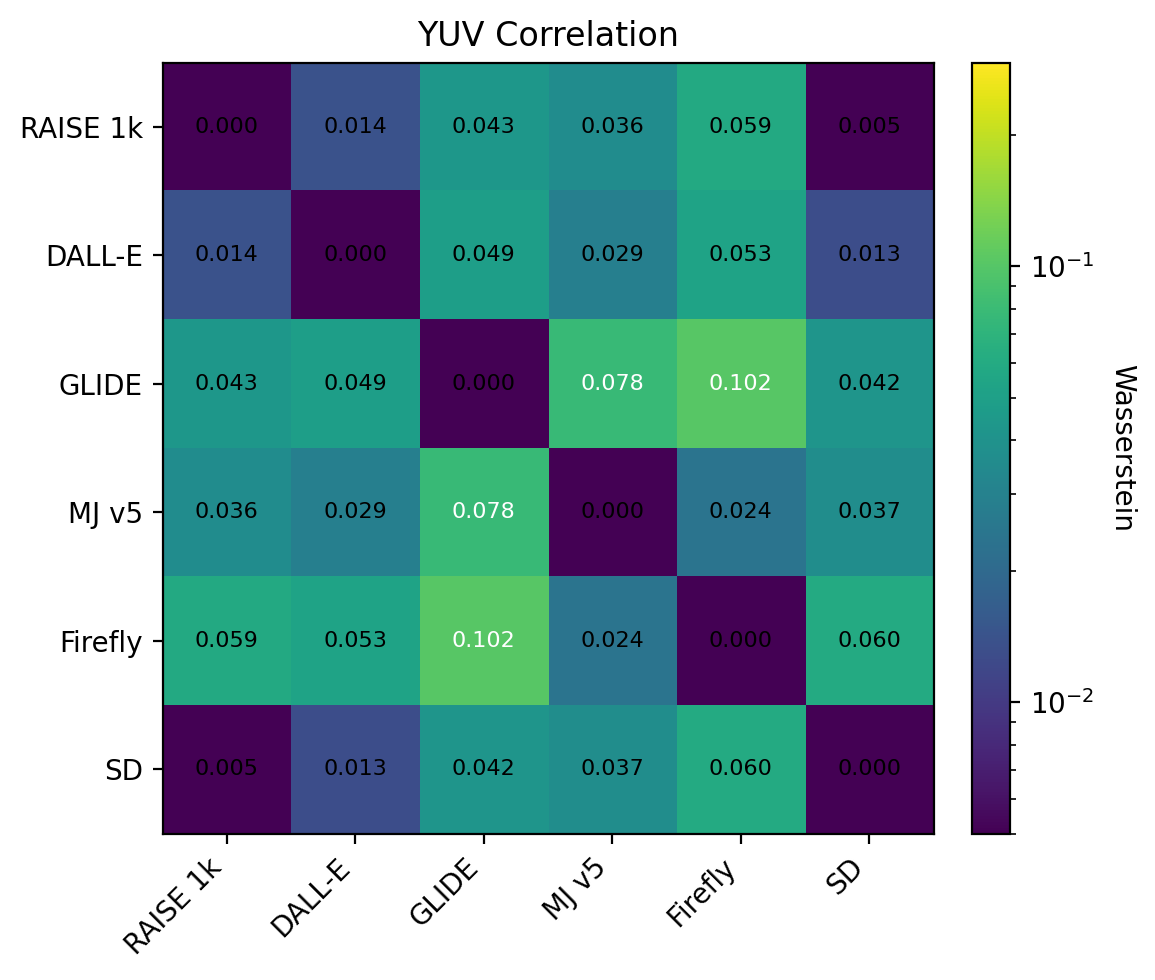}
            \vspace*{-2em}
            \caption{\hspace*{\subcapshift}YUV}
        \end{subfigure}
        \\
    \end{tabular}

    \caption{Mean Wasserstein distance matrices between inter-channel correlation feature distributions across sources, shown for multiple color spaces (RGB/HSV/Lab/YUV). Real images are from RAISE-1k~\cite{dangnguyen2015raise} and synthetic images are from Synthbuster~\cite{bammey2024synthbuster}.}
    \label{fig:corr_wasserstein_matrices}
    \vspace*{-1.2em}
\end{figure*}

Taken together, these results suggest that inter-channel correlation statistics go beyond separating real and synthetic images, encoding \emph{generator-dependent} signatures whose strength depends on the chosen color space. Since RGB and Lab provide the most consistent separation, % in both the distribution plots and the Wasserstein summaries, 
we focus on correlation features derived from these %two 
spaces in the experiments that follow.
\vspace*{-1em}

\section{From Forensic Cues to Actual Detection}
\label{sec:from_forensic_cues_to_sctual_detection}
\vspace*{-1em}

In this section, we evaluate the hypothesis that inter-channel correlation maps provide complementary evidence beyond standard RGB appearance cues, and that such cues can be exploited under a modest training budget.
To isolate the effect of the input representation, we keep the backbone fixed and compare models trained with standard RGB inputs to models augmented with inter-channel correlation maps. We consider correlation features computed in the color spaces that exhibited the strongest cross-source separation (RGB and Lab, based on the analysis in Sec.~\ref{sec:inter_channel_correlation}), both individually and in combination.

\subsection{Model and Input Representations}
\label{subsec:model_input}

\paragraph{Backbone.}\hspace{0pt}
We use ResNet-50~\cite{he2016resnet} as a simple and widely adopted CNN baseline. The network is initialized from ImageNet~\cite{deng2009imagenet} pre-training and fine-tuned for binary real-vs-generated classification with a single linear classification head.

\vspace*{-0.6em}
\paragraph{Correlation maps.}\hspace{0pt}
Let $x \in \mathbb{R}^{H\times W\times 3}$ denote an RGB image. For any color space $\mathcal{S}$ with three channels, we define $\Phi_{\mathcal{S}}(x)\in\mathbb{R}^{H\times W\times 3}$ as the stack of pairwise inter-channel correlation maps computed in $\mathcal{S}$.

\vspace*{-0.6em}
\paragraph{Input variants.}
We evaluate four input representations:
{\small
\begin{itemize}[noitemsep, topsep=0pt]
    \item \textbf{RGB}: $x \in \mathbb{R}^{H\times W\times 3}$.
    \item \textbf{RGB+Corr(RGB)}: $\mathrm{concat}\big(x,\Phi_{\mathrm{RGB}}(x)\big)\in\mathbb{R}^{H\times W\times 6}$.
    \item \textbf{RGB+Corr(Lab)}: $\mathrm{concat}\big(x,\Phi_{\mathrm{Lab}}(x)\big)\in\mathbb{R}^{H\times W\times 6}$.
    \item \textbf{RGB+Corr(RGB)+Corr(Lab)}:
    $\mathrm{concat}\big(x,\Phi_{\mathrm{RGB}}(x),\Phi_{\mathrm{Lab}}(x)\big)\in\mathbb{R}^{H\times W\times 9}$.
\end{itemize}
}

\vspace*{-0.6em}
\paragraph{Adapting the first layer.}
To accommodate $C\in\{3,6,9\}$ input channels, we replace the first convolution of ResNet-50 with a $C$-channel variant. We initialize its weights from the original RGB filters by copying/averaging them across the additional channels, while keeping all other layers unchanged. This isolates the effect of the input representation, % (RGB vs.\ correlation-augmented) 
as it does not introduce architectural changes.
\vspace*{-2em}

\subsection{Evaluation Data and Protocol}
\label{subsec:data_protocol}

We follow the benchmark protocol of~\cite{cozzolino2024raisingbar}, evaluating on its fixed test suite, spanning GAN-based, diffusion-based, and commercial text-to-image generators, and reporting per-generator AUC to quantify cross-generator generalization under a common testbed. Training uses a \emph{10k} split, which contains 10k synthetic images generated by Latent Diffusion. This setting establishes an \emph{in-domain} reference (Latent Diffusion) and enables measuring how detectors trained on a single generator generalize both to the seen source and to \emph{unseen} generators at test time.

The test set consists of 18 generators and is constructed with generator-specific real/synthetic sources: for each generator, the real images are drawn from a reference dataset associated with that generator (eg. LSUN, FFHQ), and the corresponding synthetic set is produced by the target generator under evaluation. For text-to-image systems, the protocol uses captioned real images (RAISE) and generates synthetic counterparts from the same textual descriptions, yielding semantically aligned real/fake pairs for evaluation across commercial tools.

% Motivated by Sec.~\ref{subsec:corr_distributions}, where different generators exhibit distinct correlation statistics, we also test a regime that augments the core training split of~\cite{cozzolino2024raisingbar} with a limited number of images from additional generators, as summarized in Tab.~\ref{tab:train_sources}.
% This experiment investigates whether modest multi-generator supervision enhances robustness while maintaining the backbone and training procedure unchanged.

% \begin{table}[ht]
% \centering
% \small
% \setlength{\tabcolsep}{4pt}
% \renewcommand{\arraystretch}{1.08}
% {\footnotesize
% \begin{tabular}{lcc}
% \hline
% \textbf{Source} & \textbf{\# images} & \textbf{Origin} \\
% \hline
% GLIDE & 700 & ~\cite{bammey2024synthbuster} \\
% Stable Diffusion 1.3 & 700 & ~\cite{bammey2024synthbuster} \\
% Stable Diffusion 1.4 & 700 & ~\cite{bammey2024synthbuster} \\
% Stable Diffusion 2.0 & 700 & ~\cite{bammey2024synthbuster} \\
% Stable Diffusion XL & 700 & ~\cite{bammey2024synthbuster} \\
% StyleGAN2-FFHQ & 588 &~\cite{bhargava_kaggle_deepfake_face_images} \\
% DALL{\text -}E 2 (COCO-Val) & 1000 &~\cite{cozzolino2024raisingbar} \\
% DALL{\text -}E 3 (COCO-Val) & 1000 &~\cite{cozzolino2024raisingbar} \\
% \hline
% \textbf{Core train split} & \textbf{10000} &~\cite{cozzolino2024raisingbar} \\
% \hline
% \textbf{Total} & \textbf{16088} & \\
% \hline
% \end{tabular}
% }
% \caption{Training data composition for ResNet-50 experiments.}
% \label{tab:train_sources}
% \vspace*{-3em}
% \end{table}

Motivated by Sec.~\ref{subsec:corr_distributions}, where different generators exhibit distinct correlation statistics, we also test a regime that augments the core training split of~\cite{cozzolino2024raisingbar} with a limited number of additional synthetic images from other generators, while keeping the backbone and training procedure unchanged. Specifically, we add \textbf{700} images each from \textbf{GLIDE} and five \textbf{Stable Diffusion} variants (\textbf{SD~1.3/1.4/2.0/XL}; all from Synthbuster~\cite{bammey2024synthbuster}), \textbf{588} images from \textbf{StyleGAN2-FFHQ}~\cite{bhargava_kaggle_deepfake_face_images}, and \textbf{1000} images each from \textbf{DALL-E~2} and \textbf{DALL-E~3} (COCO-Val) as in~\cite{cozzolino2024raisingbar}. Together with the \textbf{10k} core training split, this yields a total of \textbf{16,088} training images. This experiment evaluates whether modest multi-generator supervision improves cross-generator robustness by increasing coverage of generator-specific modes in correlation space.
\vspace*{-1em}

\subsection{Effect of Correlation-Augmented Inputs}
\label{subsec:ablation_discussion}
\vspace*{-1em}

Tab.~\ref{tab:input_ablation_full} evaluates whether making inter-channel correlation structure explicit provides complementary evidence for real-vs-generated classification under a fixed ResNet-50 backbone and training budget. We vary only the input representation and report per-generator AUC under the protocol of~\cite{cozzolino2024raisingbar}, considering both single-generator training and the limited multi-generator supervision regime ($\dagger$).

We observe that augmenting RGB with Lab correlation maps improves performance relative to RGB-only inputs across most generators, indicating that inter-channel dependence in a perceptual color space carries additional forensic traces.
% signal beyond appearance cues. 
However, this benefit is not universal: adding RGB correlation maps alone typically reduces AUC, and does not reliably improve when combined with Lab correlations, showing that the utility of correlation cues depends critically on the color-space parameterization.

%\paragraph{Adding generators during training helps substantially.}
In addition, results suggest that moving from single-generator training to the limited multi-generator regime ($\dagger$) yields significant gains in cross-generator robustness, particularly for correlation-augmented variants. This supports the view that correlation statistics exhibit generator-dependent modes, and that even modest exposure to additional generators improves coverage of these modes, resulting in higher AUC on unseen sources.

%\paragraph{Some generators remain challenging.}
However, despite the gains from Lab correlations and limited multi-generator supervision, several generators, especially some commercial tools, remain difficult. This suggests that within-generator source shifts and out-of-distribution processing pipelines can alter low-level statistics, including correlation structure, and that improving robustness on these cases likely requires additional domain-robust training or calibration beyond the proposed input augmentation. We hypothesize that this reflects \emph{domain mismatch and negative transfer} under the benchmark protocol: real-image sources are generator-specific, and mixing multiple generators in training can shift the decision boundary toward cues that are highly predictive for diffusion-family additions but weaker for commercial tools whose inter-channel correlation statistics appear closer to real photographs. Since these generators are black-box systems, we cannot analyze their objectives or constraints (e.g., which perceptual losses are used), limiting deeper causal interpretation.

\begin{table*}[t]
\centering
\scriptsize
\setlength{\tabcolsep}{3.0pt}
\renewcommand{\arraystretch}{1.7}
\newcommand{\gray}[1]{\textcolor{gray}{#1}}
\begin{adjustbox}{max width=\textwidth}
\begin{tabular}{l|ccccc|ccccccccc|cccc|c}
\hline
\multirow{2}{*}{\textbf{Input variant}} &
\multicolumn{5}{c|}{\textbf{GAN family}} &
\multicolumn{9}{c|}{\textbf{Diffusion family}} &
\multicolumn{4}{c|}{\textbf{Commercial tools}} &
\textbf{AVG} \\
\cline{2-19}
& \shortstack{Pro\\GAN} & \shortstack{Style\\GAN2} & \shortstack{Style\\GAN3} & \shortstack{Style\\GAN-T} & \shortstack{Giga\\GAN}
& \shortstack{Score\\SDE} & ADM & GLIDE & \shortstack{Latent\\Diff.} & \shortstack{Stable\\Diff.} & \shortstack{DeepFl.\\IF} & Ediff-I & DiT & SDXL
& \shortstack{DALL-\\E2} & \shortstack{DALL-\\E3} & Midj. & \shortstack{Adobe\\Firef.} & \\
\hline
\hline
RGB & 88.9 & 70.3 & 82.6 & 94.7 & 76.5 & 22.2 & 71.7 & 96.2 & \gray{88.2} & 99.1 & 90.5 & 73.3 & 93.9 & 93.3 & \underline{57.1} & 43.6 & 88.0 & \underline{49.0} & 76.6 \\
RGB + Corr(RGB) & 65.1 & 47.6 & 51.6 & 10.5 & 54.1 & 16.9 & 58.3 & 66.2 & \gray{62.8} & 87.5 & 79.8 & 59.3 & 70.5 & 69.9 & 42.3 & \textbf{58.1} & 49.9 & \textbf{69.2} & 56.6 \\
RGB + Corr(Lab) & 88.1 & 78.9 & 82.5 & 99.3 & 74.5 & 20.1 & 76.9 & 98.0 & \gray{93.5} & 99.6 & 92.1 & 83.5 & 96.0 & \underline{97.3} & 52.8 & 9.0 & 78.1 & 9.3 & 73.9 \\
RGB + Corr(RGB) + Corr(Lab) & 88.3 & 73.2 & 81.6 & 96.9 & 73.2 & 22.0 & 73.4 & 95.1 & \gray{89.7} & 98.7 & 90.9 & 77.3 & 94.1 & 94.2 & 52.0 & 15.5 & 83.7 & 8.8 & 72.7 \\
\hline
RGB$^{\dagger}$ & 91.0 & \gray{68.5} & 58.0 & \underline{99.9} & 77.5 & \underline{96.8} & 81.0 & \gray{\textbf{100.0}} & \gray{94.4} & \gray{\textbf{100.0}} & \underline{97.6} & 59.3 & \textbf{99.1} & \gray{\textbf{100.0}} & \gray{\textbf{100.0}} & \gray{37.1} & \textbf{99.8} & 36.9 & \underline{83.2} \\
RGB + Corr(RGB)$^{\dagger}$ & 89.4 & \gray{71.8} & 59.2 & 99.5 & 71.7 & 92.1 & 81.4 & \gray{\underline{99.8}} & \gray{93.8} & \gray{\textbf{100.0}} & \underline{97.6} & 65.8 & 94.7 & \gray{\textbf{100.0}} & \gray{\textbf{100.0}} & \gray{\underline{44.5}} & \underline{98.6} & 30.0 & 82.8 \\
RGB + Corr(Lab)$^{\dagger}$ & \underline{96.1} & \gray{\textbf{89.4}} & \underline{86.5} & \textbf{100.0} & \underline{82.7} & \textbf{97.5} & \textbf{84.8} & \gray{99.6} & \gray{\textbf{97.4}} & \gray{\underline{99.9}} & \textbf{98.8} & \textbf{87.7} & 98.3 & \gray{\textbf{100.0}} & \gray{\textbf{100.0}} & \gray{20.2} & 97.7 & 3.8 & \textbf{85.6} \\
RGB + Corr(RGB) + Corr(Lab)$^{\dagger}$ & \textbf{97.4} & \gray{\underline{85.4}} & \textbf{86.9} & \textbf{100.0} & \textbf{84.4} & 96.3 & \underline{82.5} & \gray{99.6} & \gray{\underline{95.2}} & \gray{99.8} & \textbf{98.8} & \underline{87.2} & \underline{99.0} & \gray{\textbf{100.0}} & \gray{52.0} & \gray{15.5} & 83.7 & 8.8 & 81.8 \\
\hline
\end{tabular}
\end{adjustbox}
\caption{Ablation over input representations (AUC, \%) under the benchmark protocol of~\cite{cozzolino2024raisingbar}.
$\dagger$ indicates training with additional generator data; gray marks in-domain generators.
\textbf{Bold} and \uline{underline} denote the best and second-best results per column.}
\label{tab:input_ablation_full}
\vspace*{-3em}
\end{table*}

\subsection{Comparison to Prior Detectors}
\label{subsec:sota_discussion}

Tab.~\ref{tab:sota_auc} compares our correlation-augmented ResNet-50 variants to representative detectors under the benchmark protocol of~\cite{cozzolino2024raisingbar}. While many competing approaches rely on substantially larger training budgets, specialized preprocessing, or ensembles, our objective is to quantify how far a lightweight representation change can go when paired with a standard CNN backbone.

%\paragraph{Strong in-domain performance and competitive cross-generator generalization.}
Results suggest that, on the in-domain generators associated with the training setup, \textsc{Chroma} achieves consistently high AUC, indicating that correlation-augmented inputs do not sacrifice performance on seen sources. More importantly, the method remains competitive on out-of-domain generators: across most GAN and diffusion families, our results either match or closely track the strongest baselines in the table, suggesting that Lab correlation cues capture transferable forensic evidence rather than narrow generator-specific artifacts.

%\paragraph{Implications for practical forensics.}
%The uneven behavior across generator families reinforces a common constraint in image forensics: no single cue is universally reliable across all synthesis pipelines and post-processing conditions. In practice, robust provenance decisions often benefit from combining complementary signals (e.g., spatial, spectral, and color-based evidence) and from using multiple detectors or views to reduce reliance on any single artifact channel~\cite{popescu2004statistical}. In this context, \textsc{Chroma} provides a lightweight and interpretable color-dependence cue that can be integrated with other forensic methods to improve coverage under real-world variability.

However, we also observe that \textsc{Chroma} behaves unevenly across generator families. This shows that no single forensic cue is reliable for all synthesis pipelines. While inter-channel color correlations are informative for many generators, others, such as Adobe Firefly or DALL-E3, appear to better match natural color statistics, even though they remain detectable by other methods, mainly those based on spectral artifacts~\cite{cozzolino2024raisingbar}. In practice, robust detection benefits from combining complementary cues, as different generators leave subtle but distinct imprints across representations, and only their joint consideration reveals the full forensic signature~\cite{popescu2004statistical}. In this context, \textsc{Chroma} provides a lightweight and interpretable color-dependence cue that can be integrated with other forensic methods to improve coverage under real-world variability.

%\paragraph{Scaling and data diversity.}
%While recent evidence suggests that robustness in AI-generated image detection improves primarily with \emph{generator diversity} and scale in the training set, often more than with architectural complexity alone~\cite{park2025communityforensics}. 
%From a complementary perspective, 
Relatedly, recent work shows that robustness in synthetic image detection is driven %largely 
by generator diversity and training scale, often more so than by architectural complexity~\cite{park2025communityforensics}.
In this context, it is notable that \textsc{Chroma} attains competitive cross-generator performance using a standard ResNet-50 and a comparatively small training split, indicating that correlation cues can be learned efficiently under limited supervision. This also suggests a clear path forward: coupling correlation-augmented inputs with larger and more diverse synthetic corpora could further improve coverage of generator-specific modes in correlation space and mitigate the failure cases observed on some commercial tools.

\begin{table*}[t]
\centering
\scriptsize
\setlength{\tabcolsep}{3.0pt}
\renewcommand{\arraystretch}{1.7}
\newcommand{\gray}[1]{\textcolor{gray}{#1}}
\begin{adjustbox}{max width=\textwidth}
\begin{tabular}{l|ccccc|ccccccccc|cccc|c}
\hline
\multirow{2}{*}{\textbf{Method}} &
\multicolumn{5}{c|}{\textbf{GAN family}} &
\multicolumn{9}{c|}{\textbf{Diffusion family}} &
\multicolumn{4}{c|}{\textbf{Commercial tools}} &
\textbf{AVG} \\
\cline{2-19}
& \shortstack{Pro\\GAN} & \shortstack{Style\\GAN2} & \shortstack{Style\\GAN3} & \shortstack{Style\\GAN-T} & \shortstack{Giga\\GAN}
& \shortstack{Score\\SDE} & ADM & GLIDE & \shortstack{Latent\\Diff.} & \shortstack{Stable\\Diff.} & \shortstack{DeepFl.\\IF} & Ediff-I & DiT & SDXL
& \shortstack{DALL-\\E2} & \shortstack{DALL-\\E3} & Midj. & \shortstack{Adobe\\Firef.} & \\
\hline
Wang et al.  & \gray{100.0} & \underline{96.5} & \underline{98.5} & 98.9 & 66.6 & 32.9 & 64.3 & 48.5 & 59.2 & 41.5 & 78.0 & 64.9 & 58.6 & 54.3 & 64.8 & 10.9 & 40.2 & 84.8 & 64.6 \\
PatchFor.    & 92.3 & 84.5 & 91.8 & 91.2 & 64.7 & 83.3 & 74.8 & \underline{96.2} & 78.1 & 62.4 & 62.7 & 78.7 & 83.1 & 68.4 & 41.9 & 52.7 & 57.8 & 49.4 & 73.0 \\
Grag. et al. & \gray{100.0} & \textbf{99.8} & 97.5 & 98.8 & 82.8 & 92.1 & 74.7 & 62.8 & \underline{91.9} & 52.5 & 69.9 & 69.6 & 65.3 & 58.0 & 58.3 & 2.4  & 43.1 & 63.5 & 71.3 \\
Mand. et al. & \underline{96.2} & \gray{93.8} & \textbf{100.0} & 92.6 & 61.8 & \gray{99.8} & 56.5 & 40.5 & 70.0 & 36.8 & 47.2 & 65.0 & 59.1 & 27.0 & 14.5 & 14.7 & 24.3 & 36.7 & 57.6 \\
Liu et al.   & \gray{100.0} & \textbf{99.8} & 98.4 & 98.5 & \textbf{98.2} & 95.4 & 82.5 & 76.5 & \textbf{97.6} & 77.4 & 72.2 & 98.7 & 88.0 & 31.1 & 70.4 & 0.2  & 40.7 & 11.8 & 74.3 \\
Corvi et al. & 79.4 & 73.7 & 50.0 & 97.1 & 63.4 & 65.0 & 80.7 & 91.9 & \gray{100.0} & \textbf{100.0} & \textbf{99.9} & 85.7 & \textbf{100.0} & \textbf{100.0} & 69.4 & \underline{60.8} & \textbf{100.0} & \textbf{98.0} & 84.2 \\
LGrad        & \gray{100.0} & 91.2 & 83.8 & 81.8 & 82.2 & 80.6 & 76.9 & 66.1 & 81.1 & 61.5 & 68.8 & 74.1 & 56.2 & 57.2 & 58.6 & 37.9 & 56.3 & 40.6 & 69.7 \\
Ojha et al.  & \gray{100.0} & 93.9 & 92.3 & 98.2 & \underline{96.0} & 58.4 & \textbf{86.7} & 80.8 & 85.7 & 89.5 & 92.9 & 80.6 & 77.8 & 85.1 & \textbf{95.2} & 36.4 & 66.2 & \underline{97.5} & 84.1 \\
DIRE-1       & 50.6 & 56.9 & 47.8 & \underline{99.9} & 74.1 & 44.3 & \gray{75.7} & 71.4 & 68.7 & 39.4 & \textbf{98.9} & \underline{99.1} & \underline{99.6} & 47.1 & 44.7 & 47.6 & 51.0 & 57.4 & 65.2 \\
DIRE-2       & 54.2 & 52.5 & 43.0 & 99.6 & 76.0 & 41.0 & 70.1 & 70.1 & 69.3 & 46.9 & 97.0 & 98.2 & 98.3 & 42.8 & 41.0 & 49.6 & 47.8 & 43.0 & 63.3 \\
NPR          & \gray{100.0} & 85.6 & 77.0 & 96.4 & 88.7 & 91.1 & \underline{86.3} & 79.3 & 90.2 & 64.5 & 91.6 & 80.1 & 78.4 & 76.7 & 39.5 & 48.7 & 77.0 & 32.1 & 76.8 \\
Cozzolino et al. 10k+ & 93.4 & 87.1 & 87.6 & \underline{99.9} & 78.5 & 89.2 & 79.9 & \textbf{99.7} & \gray{84.7} & \underline{91.3} & 97.9 & \textbf{99.4} & 94.0 & \underline{90.1} & \underline{86.3} & \textbf{92.9} & 81.7 & 87.2 & \textbf{90.0} \\
\hline
RGB+Corr(Lab)$^{\dagger}$ & 96.1 & \gray{89.4} & 86.5 & \textbf{100.0} & 82.7 & \textbf{97.5} & 84.8 & \gray{99.6} & \gray{97.4} & \gray{99.9} & 98.8 & 87.7 & 98.3 & \gray{\textbf{100.0}} & \gray{100.0} & \gray{20.2} & \underline{97.7} & 3.8 & \underline{85.6} \\
RGB+Corr(RGB)+Corr(Lab)$^{\dagger}$ & \textbf{97.4} & \gray{85.4} & 86.9 & \textbf{100.0} & 84.4 & \underline{96.3} & 82.5 & \gray{99.6} & \gray{95.2} & \gray{99.8} & 98.8 & 87.2 & 99.0 & \gray{\textbf{100.0}} & \gray{52.0} & \gray{15.5} & 83.7 & 8.8 & 81.8 \\
\hline
\end{tabular}
\end{adjustbox}
\caption{Comparison with prior methods in terms of AUC (\%) under the benchmark protocol of~\cite{cozzolino2024raisingbar}. Results are reported across GAN-based generators, diffusion-based generators, and commercial tools. Gray entries indicate the in-domain generator(s) associated with each method’s training setup as reported by the protocol, whereas the remaining columns measure cross-source generalization. \textbf{Bold} marks the best overall average and \uline{underline} marks the second-best.}
\label{tab:sota_auc}
\vspace*{-2.5em}
\end{table*}

\section{Conclusions}
\label{sec:conclusion}

This work investigated \emph{inter-channel color-space correlations} as a lightweight and underexploited forensic cue for AI-generated image detection. Motivated by the observation that a standard perceptual metric (LPIPS) exhibits non-uniform sensitivity to perturbations that selectively alter cross-channel dependence across different color parameterizations, we analyzed local Pearson correlation features across multiple color spaces on semantically matched real and synthetic data. The resulting distributional study revealed systematic, generator-dependent shifts in correlation space, with RGB and, especially, Lab providing the clearest separation between real photographs and generated imagery.

Building on these findings, we introduced \textsc{Chroma}, a simple correlation-augmented representation that concatenates standard RGB inputs with correlation maps, and evaluated it using a fixed ResNet-50 backbone under a modest training budget. Experiments showed that explicitly providing \emph{Lab} correlation structure yields consistent gains in real-vs-generated discrimination and cross-generator robustness over RGB-only training, supporting the claim that cross-channel dependencies encode complementary forensic information beyond appearance cues. Importantly, these gains are \emph{representation-dependent}: correlation maps computed directly in RGB are not reliably beneficial and can degrade performance, highlighting that the choice of color-space parameterization is central to the utility of correlation-based cues.

More broadly, our results reinforce recent evidence that robustness improves with exposure to \emph{generator diversity} in training data. Despite using a comparatively small training split and a standard CNN backbone, \textsc{Chroma} attains competitive cross-generator performance across most GAN and diffusion families, suggesting that explicit correlation cues can be learned efficiently under limited supervision and may benefit further from scaling training diversity. At the same time, several commercial tools remain challenging, consistent with within-generator source shifts and out-of-distribution post-processing pipelines that can alter low-level statistics and reduce separability.

Several directions remain for future work. First, correlation cues could be integrated with complementary forensic traces %(e.g., spectral residuals) 
in a unified multi-view model to improve robustness. %to post-processing. 
%Second, one could learn a non-linear ``forensic'' color space (e.g., via an MLP or $1{\times}1$ convolutions) and compute correlation maps in that learned space. 
Second, rather than relying on fixed color spaces, one could learn a non-linear forensic color space (e.g., via an MLP or $1{\times}1$ convolutions) in which discrimination is maximized. Finally, extending correlation-based cues to localization tasks (e.g., inpainting) and to detector calibration under domain shift may further increase their utility in broader forensic workflows.

\bibliographystyle{splncs04}
\bibliography{bibliography}

\end{document}